\documentclass[journal]{IEEEtran}

\ifCLASSINFOpdf

\else

\fi

\usepackage{graphicx,wrapfig,fullpage,amsmath,hhline,epsfig,verbatim,url,amssymb,multicol,multirow,cite}
\usepackage{times,color,soul} 
\usepackage{amsthm,amssymb}
\usepackage[left=0.75in,top=0.70in,right=0.75in,bottom=0.80in]{geometry}
\usepackage{amsbsy,latexsym,mathrsfs,mathtools}
\usepackage{mathptmx}
\DeclareMathAlphabet{\mathcal}{OMS}{cmsy}{m}{n}
\usepackage{bm}
\usepackage{float}
\usepackage{color,soul}
\usepackage{dsfont}
\usepackage{comment}
\usepackage{algorithm}
\usepackage{algorithmic}
\usepackage{psfrag}   
\usepackage{tabularx}

\usepackage{url}
\usepackage{hyperref}
\usepackage{setspace}
\usepackage[table]{xcolor}
\usepackage{paralist}
\usepackage{booktabs}
\usepackage{gensymb}
\usepackage{flushend}

\newcommand{\YanMod}[1]{\textcolor{black}{#1}}
\newcommand{\AmirMod}[1]{\textcolor{black}{#1}}

\usepackage[switch]{lineno}

\begin{document}

\title{\YanMod{Real-Time Walking Pattern Generation of Quadrupedal Dynamic-Surface Locomotion based on a Linear Time-Varying Pendulum Model}}

\author{Amir~Iqbal$^{1}$,~
        Sushant~Veer$^{2}$, and~
        Yan~Gu$^{3,\dagger}$
\thanks{$^{1}$A. Iqbal is with the Department of Mechanical Engineering, University
of Massachusetts Lowell, Lowell, MA 01854, U.S.A.
{\tt\small amir\_iqbal@student.uml.edu.}
$^{2}$S. Veer is with NVIDIA Research, Santa Clara, CA 95051, U.S.A. This work was conducted while S. Veer was with Princeton University.
{\tt\small sveer@nvidia.com.}
$^{3}$Y. Gu is with the School of Mechanical Engineering, Purdue University, West Lafayette, IN 47907, U.S.A.
{\tt\small yangu@purdue.edu.}
Corresponding author: Y. Gu.}
}

\maketitle

\begin{abstract}
\YanMod{This study introduces an analytically tractable and computationally efficient model of the legged robot dynamics associated with locomotion on a dynamic rigid surface (DRS), and develops a real-time motion planner based on the proposed model and its analytical solution. 
This study first theoretically extends the classical linear inverted pendulum (LIP) model from legged locomotion on a static surface to DRS locomotion, by relaxing the LIP's underlying assumption that the surface is static.
The resulting model, which we call ``DRS-LIP'', is explicitly time-varying.
After converting the DRS-LIP into Mathieu's equation, an approximate analytical solution of the DRS-LIP is obtained, which is reasonably accurate with a low computational cost.
Furthermore, to illustrate the practical uses of the analytical results, they are exploited to develop a hierarchical motion planner that efficiently generates physically feasible trajectories for DRS locomotion.
Finally, the effectiveness of the proposed theoretical results and motion planner is demonstrated both through PyBullet simulations and experimentally on a Laikago quadrupedal robot that walks on a rocking treadmill.}
\AmirMod{The videos of simulations and hardware experiments are available at \href{https://youtu.be/u2Q_u2pR99c}{\tt \small https://youtu.be/u2Q\_u2pR99c.}}

\end{abstract}

\begin{IEEEkeywords}
Legged locomotion, nonstationary surfaces, dynamic modeling, analytical solution, motion planning.
\end{IEEEkeywords}

\IEEEpeerreviewmaketitle

\vspace{-0.15 in}
\section{Introduction}
\vspace{-0.05 in}
Legged robots have the potential to traverse various challenging surfaces, including stationary (uneven or discrete) surfaces~\cite{anymal_16,fawcett2021robust,zhang2021efficient, mastalli2020motion,shin2022design,gu2018exponential,gao2019global} and nonstationary rigid surfaces (i.e., rigid surfaces that move in the inertial frame)~\cite{iqbal2020provably,gao2021invariant}.
Legged robots capable of reliably traversing a dynamic rigid surface (DRS) can aid in various critical real-world applications such as firefighting, maintenance, and inspection on ships and public transit vehicles.
The objective of this study is to model and analyze the essential dynamic behaviors of a legged robot that walks on a DRS,
\YanMod{and to exploit these analytical results for efficient motion planning of legged locomotion}.
\YanMod{There has been ample work on reduced-order modeling and motion planning of legged locomotion on stationary surfaces, but not for DRS.
This paper constitutes one of the first attempts to build a reduced-order model and leverage such a model in motion planning for DRS locomotion.}
Yet, reduced-order modeling and planning of DRS locomotion is fundamentally complex due to the nonlinear robot dynamics~\cite{shih2012stable, motahar2016composing,gao2019global2} and the time-varying movement of surface-foot contact points~\cite{iqbal2020provably,gao2022invariant}.

\vspace{-0.1 in}
\subsection{\YanMod{Reduced-Order Models of Legged Locomotion on Stationary or Dynamic Surfaces}}
\vspace{-0.05 in}

A reduced-order dynamics model of legged locomotion captures the robot's essential dynamic behaviors~\cite{chen2020optimal}.
One of the most widely studied reduced-order models for stationary surface walking is the linear inverted pendulum (LIP) model \cite{kajita20013d}, which approximates a legged robot as a point mass atop a massless leg.
Many of today's walking robots can be relatively accurately modeled as the LIP since they typically have a heavy upper body and lightweight legs~\cite{pratt2006capture,mastalli2020motion}.

Due to its simplicity, the LIP is analytically tractable and can provide physical insights into the essential robot dynamics. 
It also explicitly reveals the simplified relationship between the center of pressure (CoP), which can be used to infer the feasibility of ground contact forces (i.e., no foot rolling about any edge of the region of contact), and the center of mass (CoM).
Thus, the LIP can serve as a basis of motion planning for ensuring the computational efficiency and physical feasibility of planning, \YanMod{as reviewed later.}

The classical LIP~\cite{kajita20013d} \YanMod{for static surfaces} has been extended to various complex scenarios such as foot sliappge~\cite{mihalec2020recoverability}, a varying CoM height~\cite{Cap_uneve_caron2019TRO}, CoM motions on 3-D planes~\cite{zhao2017robust}, nontrivial centroidal angular momentum~\cite{pratt2006capture}, and hybrid robot dynamics~\cite{xiong2022_3d,dai2022data}.
Due to their static surface assumption, they may not be suitable for DRSes with significant motions.

For locomotion on a DRS whose motions are affected by the robot (e.g., passive and relatively lightweight surfaces), several reduced-order robot dynamics models have been recently introduced, including extended LIP~\cite{Yamenzheng2011ball}, centroidal dynamics~\cite{BallMan2020Koshil_yang}, and rimless-wheel models~\cite{asano2021modeling}.
Still, it is unclear how to extend these models to DRSes whose motion cannot be affected by the robot (e.g., trains, vessels, and elevators).
For such substantially heavy or rigidly actuated DRSes, the effects of the DRS motion on a spring-loaded inverted pendulum model have been numerically studied~\cite{iqbal_SLIP}.
\YanMod{However, the stability conditions and analytical solution of the model remain unknown.}

Beyond the scope of legged locomotion, the modeling and analysis of an inverted pendulum with a vertically oscillating support, i.e., the Kapitza pendulum~\cite{kapitza1951dynamic}, is a classical physics problem.
The Kapitza pendulum has an intriguing property that under high-frequency support oscillations, the pendulum's upper equilibrium becomes stable whereas its lower one is unstable.
Yet, it is an open question whether and when the Kapitza pendulum is a reasonable approximation of DRS locomotion.
Also, the motion frequencies of real-world DRSes (e.g., vessels~\cite{ShipMotion_tannuri2003estimating}) are commonly too low to meet the conditions underlying the pendulum.

\vspace{-0.15 in}
\subsection{\YanMod{Motion Planning based on Inverted Pendulum Models}}
\vspace{-0.05 in}

\YanMod{Since the LIP model represents the low-dimensional CoM dynamics of robot walking, it has been utilized to efficiently plan physically feasible walking motions on a static surface.}
\YanMod{Given the user-specified footstep and CoP positions,
the exact closed-form analytical solution of the classical LIP~\cite{Kajita2003BipedWP,gong2020angular} 
has been used to enable real-time planning of feasible CoM trajectories for static surface walking.
This analytical solution has been augmented with the discrete-time jump of the CoM position (relative to the CoP) at a foot-landing event, which is then used to derive the desired footstep locations that provably stabilize the hybrid LIP model \cite{xiong2022_3d}.
Recently, the exact capturability conditions of a LIP model with a time-varying CoM height have been derived based on the closed-form solution of the model's time-varying damping function at a robot's desired final CoM state~ \cite{Cap_uneve_caron2019TRO}. 
These conditions are then used to plan the desired CoM and CoP trajectories with provable capturability guarantees.
As reviewed earlier, the underlying LIP models of these planners assume a stationary walking surface, and thus the planner may not be directly used for DRS locomotion.}

\vspace{-0.15 in}
\subsection{Contributions}
\vspace{-0.05 in}

\AmirMod{This study aims to theoretically extend the classical LIP model~\cite{kajita20013d} from stationary surfaces to substantially heavy or rigidly actuated DRSes (e.g., ships), introduce an analytical approximate solution to the extended LIP model (termed as ``DRS-LIP"), and develop and experimentally validate a real-time motion planner that uses the proposed solution to ensure planning efficiency and feasibility.
A preliminary version of this work appeared in \cite{iqbal2021extended} where we derived the DRS-LIP model.
The new, substantial contributions of this study compared to \cite{iqbal2021extended} are:}
\AmirMod{
\begin{enumerate}
    \item [(a)] Forming the analytical approximate solution of the DRS-LIP under a vertical, sinusoidal DRS motion and giving physical insights into the model's stability, which are both missing in \cite{iqbal2021extended}.
     \item [(b)] Assessing the accuracy and computational efficiency of the proposed analytical approximate solution through comparison with a highly accurate numerical solution in MATLAB, which is not included in \cite{iqbal2021extended}.
     \item [(c)] Designing a hierarchical walking pattern generator that utilizes the proposed analytical solution to efficiently plan feasible robot motions, whereas the previous reduced-order model based planner in \cite{iqbal2021extended} does not explicitly ensure the feasibility of the generated motion.
     \item [(d)] Validating the planner efficiency and feasibility through both realistic PyBullet simulations and hardware experiments under various surface and robot movements, while no \emph{hardware experiment} results are given in \cite{iqbal2021extended}.
\end{enumerate}
}

\vspace{-0.1 in}
\section{Reduced-Order Model of DRS Locomotion}
\vspace{-0.05 in}
\label{Modeling}
\label{sec: model}

This section introduces a reduced-order model that captures the essential robot dynamics associated with legged walking on a DRS.
The model is derived by extending the classical LIP model~\cite{kajita20013d} from static surfaces to a DRS, and is called ``DRS-LIP''.

Today's legged robots typically have a heavy upper body and lightweight legs.
Their CoM dynamics can be approximately described by a LIP, i.e., a point mass atop a massless leg~\cite{kajita20013d}, under the assumption that:
\begin{itemize} 
    \item [(A1)] The robot's rate of whole-body angular momentum about the CoM is negligible.
\end{itemize}
Assumption (A1) is reasonable for real-world locomotion because the robot's trunk is typically controlled to maintain a steady orientation for housing sensors (e.g., cameras).

\begin{figure}[t]
    \centering
    \includegraphics[width= 0.9\linewidth]{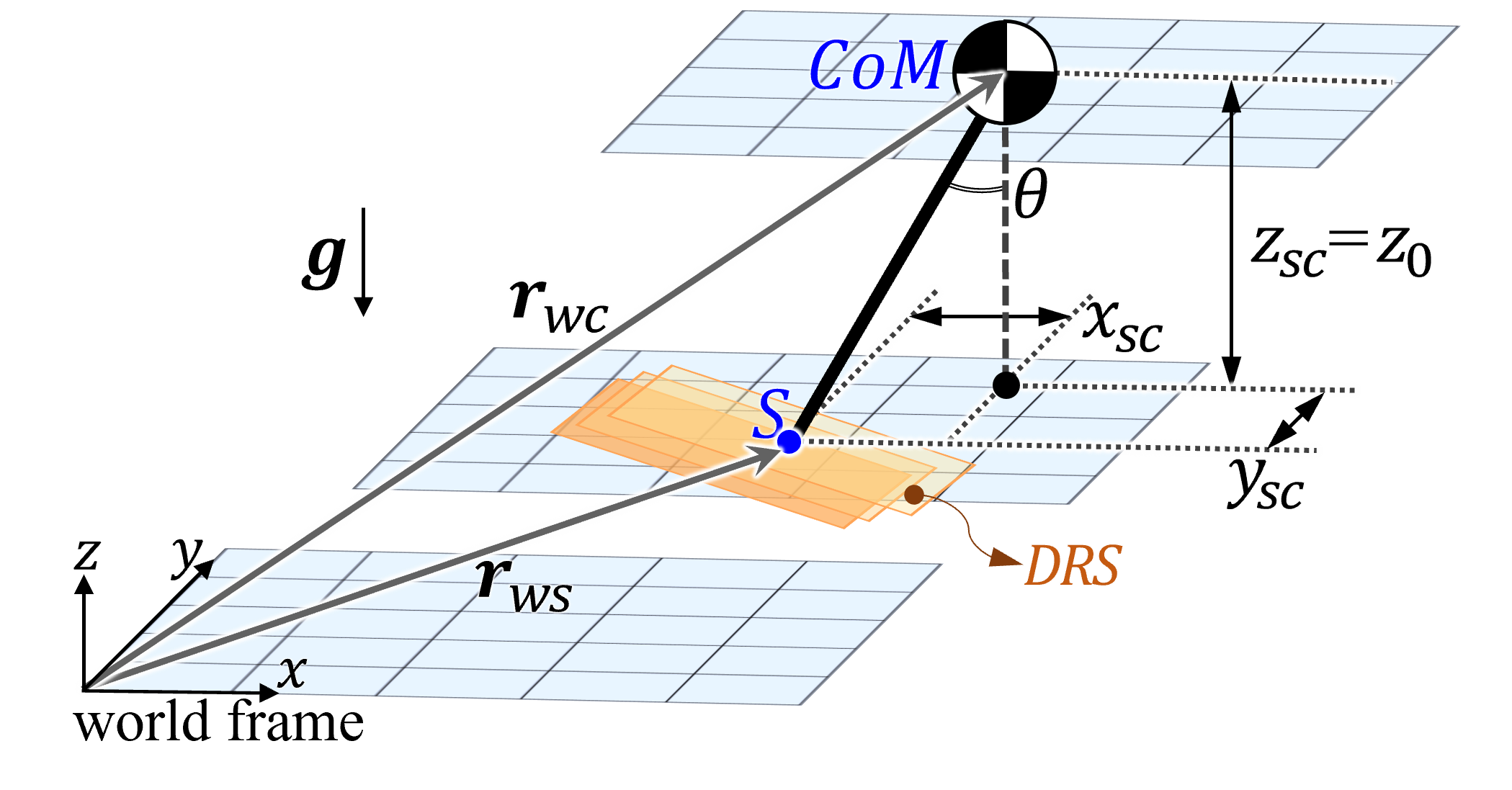}
    \vspace{-0.2 in}
    \caption{\YanMod{Illustration of the proposed DRS-LIP model.
    All three grid planes are horizontal. 
    The top and middle ones pass through the CoM and the leg's far end $S$, respectively.
    The bottom one is fixed to the world frame.}}
    \label{Fig:LIPM_sketch}
    \vspace{-0.2 in}
\end{figure}

In this study, we use a 3-D LIP to capture the essential dynamics of a 3-D legged robot walking on a DRS (see Fig.~\ref{Fig:LIPM_sketch}).
The point mass and support point {\small $S$} in Fig.~\ref{Fig:LIPM_sketch} correspond to the robot's CoM and CoP.

Let {\small $\mathbf{r}_{wc}=[x_{wc},~y_{wc},~z_{wc}]^T$} and {\small $\mathbf{r}_{ws}=[x_{ws},~y_{ws},~z_{ws}]^T$} respectively denote the positions of the CoM and point {\small $S$} in the world frame.
Then, the CoM position relative to point $S$, denoted as {\small $\mathbf{r}_{sc}$}, is defined as: {\small $\mathbf{r}_{sc}=\mathbf{r}_{wc} -\mathbf{r}_{ws}=:[x_{sc},~y_{sc},~z_{sc}]^T$}.

The CoM dynamics during DRS locomotion are given by:
\begin{equation}
\small
    \ddot{x}_{wc} = \frac{f_a x_{sc}}{m r} \sin \theta,
    ~ \ddot{y}_{wc} = \frac{f_a y_{sc}}{m r} \sin \theta ,
    ~
    \ddot{z}_{wc} = \frac{f_a}{m} \cos \theta -g.
     \label{Eq: LIPM_EOM_world - z}
\end{equation}
Here, {\small $m$} is the robot's total mass,
{\small $\theta$} is the angle of {\small $\mathbf{r}_{sc}$} relative to the vertical axis,
{\small $g$} is the norm of the gravitational acceleration,
{\small $r$} is the projected length of {\small $\mathbf{r}_{sc}$} on the horizontal plane,
and {\small $f_a$} is the norm of the ground contact force pointing from point {\small $S$} to the CoM. 

\vspace{-0.15 in}
\subsection{DRS-LIP under a General Vertical Surface Motion}
\vspace{-0.05 in}

We consider the following assumption on the vertical distance {\small $z_{sc}$} between the CoM and point {\small $S$} (see Fig.~\ref{Fig:LIPM_sketch}):
\begin{itemize} 
    \item[(A2)] The CoM maintains a constant height $z_0$ above the support point {\small $S$} (i.e., {\small $z_{sc} = z_0$}).
\end{itemize}
This assumption is analogous to the simplifying assumption of the classical LIP model that the point-mass height over the stationary surface is constant~\cite{kajita20013d}.

Under assumption (A2), the relationships
{\small $\dot{z}_{wc} = \dot{z}_{ws}$} and {\small $\ddot{z}_{wc} = \ddot{z}_{ws}$} hold, and then the axial force {\small $f_a$} becomes
{\small $ f_a = m (\ddot{z}_{ws}+g)/\cos \theta$}.
Thus, the horizontal LIP dynamics are:
\begin{equation}
\small
     \ddot{x}_{wc} =
     (\ddot{z}_{ws}+g)\frac{x_{sc}}{z_0}
     ~\text{and}~
    \ddot{y}_{wc}= 
    (\ddot{z}_{ws}+g)\frac{y_{sc}}{z_0}.
     \label{Eq:XY-motion general}
\end{equation}
Then, by substituting {\small $\ddot{x}_{wc}=\ddot{x}_{ws}+ \ddot{x}_{sc}$} and {\small $\ddot{y}_{wc}=\ddot{y}_{ws}+ \ddot{y}_{sc}$} into \eqref{Eq:XY-motion general}, the horizontal LIP dynamics become:
\begin{equation}
\small
    \ddot{x}_{sc}  - \frac{(\ddot{z}_{ws}+g)}{z_0} x_{sc} = -\ddot{x}_{ws}
   ~ \text{and}~
    \ddot{y}_{sc}  - \frac{(\ddot{z}_{ws}+g)}{z_0} y_{sc} = -\ddot{y}_{ws}.
    \label{Eq:simplified_Cap-xy}
\end{equation}
When there is no slippage between the support point {\small $S$} and the surface, the acceleration of point {\small $S$}, {\small $(\ddot{x}_{ws},\ddot{y}_{ws},\ddot{z}_{ws})$}, equals the DRS' acceleration at {\small $S$}.
Given that real-world DRSes (e.g., vessels) are typically equipped with high-accuracy, real-time motion monitoring systems~\cite{Ship_Motion_Monitoring_System}, we assume the time profile of {\small $(\ddot{x}_{sw},\ddot{y}_{sw},\ddot{z}_{sw})$} is known.
Accordingly, they are treated as explicit time functions.
Thus, the dynamics in~\eqref{Eq:simplified_Cap-xy} are linear, nonhomogeneous, and time-varying.

Since DRSes, such as cruising ships in regular sea waves, have relatively small horizontal acceleration compared with vertical acceleration~\cite{gahlinger2000const_hv_ship,Vessels_Vertical_Periodic_1954,gao2022invariant},
we assume the horizontal acceleration of point {\small $S$} is sufficiently small to be ignored:
\begin{itemize} 
    \item[(A3)] The horizontal accelerations of point {\small $S$} (i.e., {\small $\ddot{x}_{ws}$} and {\small $\ddot{y}_{ws}$}) are negligible.
\end{itemize}
Then, the forcing terms in \eqref{Eq:simplified_Cap-xy} (i.e., {\small $-\ddot{x}_{ws}$} and {\small $-\ddot{y}_{ws}$}) can be approximated as zero, and the horizontal LIP dynamics in \eqref{Eq:simplified_Cap-xy} become linear, time-varying, and homogeneous:
\begin{equation}
\small
    \ddot{x}_{sc}  - \frac{(\ddot{z}_{ws}+g)}{z_0} x_{sc} = 0
    ~\text{and}~
    \ddot{y}_{sc}  - \frac{(\ddot{z}_{ws}+g)}{z_0} y_{sc} = 0.
    \label{Eq-LIPM_on_DRS_simplified} 
\end{equation}
Note that the vertical CoM trajectory is given by: {\small $z_{sc} = z_0$}.

\noindent \textbf{Remark 1 \YanMod{(DRS-LIP)}:}
The LIP model in \eqref{Eq-LIPM_on_DRS_simplified}, along with {\small $z_{sc} = z_0$}, describes the simplified dynamics of DRS walking under assumptions (A1)-(A3), which we call ``DRS-LIP''.

\vspace{-0.12 in}
\subsection{DRS-LIP under a Vertical Sinusoidal Surface Motion}
\vspace{-0.05 in}

A real-world DRS, such as a vessel in regular sea waves, typically exhibits a vertical, sinusoidal motion with a constant amplitude and frequency~\cite{Vessels_Vertical_Periodic_1954}.
Thus, we focus on such motions for further analysis of the DRS-LIP. 

Under a vertical, sinusoidal surface motion, the vertical acceleration {\small $\ddot{z}_{ws}$} of point {\small $S$} is sinusoidal, and \eqref{Eq-LIPM_on_DRS_simplified} becomes the well-known Mathieu's equation~\cite{farkas2013periodic}, as explained next.

Without loss of generality, the vertical sinusoidal motion of the DRS at the surface-foot contact point is assumed as:
\begin{equation}
\small
    z_{ws} = A \sin \omega t,
    \label{Eq-ref: sinusoidal DRS motion}
\end{equation}
where the real scalar parameters {\small $A$} and {\small $\omega$} are the amplitude and frequency of the vertical surface motion, respectively.

Then, the surface acceleration {\small $\ddot{z}_{ws}$} at the support point is {\small $\ddot{z}_{ws} :=-A \omega^2 \sin \omega t$}, with which \eqref{Eq-LIPM_on_DRS_simplified} becomes:
\begin{equation}
        \ddot{x}_{sc}  - \tfrac{(g-A \omega^2 \sin \omega t)}{z_0} x_{sc} = 0
        ~ \text{and}~         
        \ddot{y}_{sc}  - \tfrac{(g-A \omega^2 \sin \omega t)}{z_0} y_{sc} = 0.
        \label{Eq-Hills_vs_motion}
\end{equation}
In \eqref{Eq-Hills_vs_motion} the two equations in the {\small $x$}- and {\small $y$}-directions are decoupled and share the same structure. Thus, their solutions share the same form.
For brevity, we focus on deriving the solution along the $x$-direction, {\small ${x}_{sc}$}, in Sec.~\ref{sec: solution}.

With a new time variable {\small $\tau :=\frac{{\pi}+2\omega t}{4}$},
the DRS-LIP in~\eqref{Eq-Hills_vs_motion} can be transformed into the standard
Mathieu's equation as:
\begin{equation}
\small
    \frac{d^2x_{sc}}{d\tau^2} +(c_0 - 2c_1\cos 2 \tau)x_{sc}  = 0,
    \label{Eq-transformed_MathieuEqn}
\end{equation}
where the real scalar coefficients {\small $c_0$} and {\small $c_1$} are defined as {\small $c_0 := -\frac{4g}{\omega^2 z_0}$} and {\small $c_1 := \frac{2A}{z_0}$}.

\vspace{-0.1 in}
\section{APPROXIMATE ANALYTICAL SOLUTION}
\label{sec: solution}

This section introduces an approximate analytical solution of the DRS-LIP under a vertical, sinusoidal DRS motion.

\vspace{-0.1 in}
\subsection{Approximation of Exact Analytical Solution}
\vspace{-0.05 in}

The DRS-LIP model in~\eqref{Eq-transformed_MathieuEqn} generally does not have an exact, closed-form analytical solution.
One straightforward approach to derive an approximate analytical solution is to utilize the fundamental solution matrix based on the Floquet theory~\cite{farkas2013periodic}.
Alternatively, we choose to exploit the existing analytical results of the well-studied Mathieu's equation to obtain a more computationally efficient solution.

There are various existing analytical approximate solutions 
of Mathieu's equation, including periodic solutions~\cite{phelps1965analytical} and those expressed through power series~\cite{farkas2013periodic}.
In this study, we adopt the general, exact analytical solution from~\cite{BookB_ME_werth2005charged} because of its generality and computational efficiency:

\noindent \textbf{Theorem 1 \YanMod{(Exact solution of Mathieu's equation~\cite{BookB_ME_werth2005charged})}:}
{\it The exact, general (periodic or non-periodic) analytical solution of Mathieu's equation in~\eqref{Eq-transformed_MathieuEqn} is as follows:}
\begin{equation}
\small
    {x}_{sc}(\tau) = \alpha_1 e^{\mu \tau}\sum_{n=-\infty}^{\infty}C_{2n} e^{i2n\tau} +\alpha_2 e^{-\mu \tau}\sum_{n=-\infty}^{\infty}C_{2n} e^{-i2n\tau}.
    \label{Eq-Assumed_GenSol_of_MathieuEqn}
\end{equation}
{\it Here, {\small $\mu$} is the characteristic exponent of~\eqref{Eq-transformed_MathieuEqn}.
{\small $\alpha_1$} and {\small $\alpha_2$} are real scalar coefficients,
{\small $n$} is an integer,
\YanMod{{\small $i$} is a unit imaginary number,}
and {\small $C_{2n}$}'s are complex scalar coefficients.}

The proof of Theorem 1 can be readily obtained based on~\cite{BookB_ME_werth2005charged}.
To use~\eqref{Eq-Assumed_GenSol_of_MathieuEqn} to compute an approximate solution,
we need to determine the number of terms to keep in the approximate solution as well as the values of the parameters {\small $\mu$}, {\small $\alpha_1$}, {\small $\alpha_2$}, and {\small $C_{2n}$}'s, which is explained next.

\subsubsection{Obtaining characteristic exponent {\small $\mu$}}
Substituting the exact solution \eqref{Eq-Assumed_GenSol_of_MathieuEqn} into \eqref{Eq-transformed_MathieuEqn} yields a recurrence relationship:
\begin{equation}
\small
    \beta_n(\mu)C_{2(n-1)} + C_{2n} + \beta_n(\mu)C_{2(n+1)} =0,
    \label{Eq: recurrence relation}
\end{equation}
where the complex scalar function {\small $\beta_n$} is {\small $\beta_n(\mu):=\frac{c_1}{(2n-i\mu)^2-c_0}.$}
The derivation of \eqref{Eq: recurrence relation} is given in \AmirMod{Appendix~A} and \cite{BookB_ME_werth2005charged}.

Equation \eqref{Eq: recurrence relation} for all {\small $n\in \mathbb{Z}^+$}
generates the following infinite set of linear homogeneous equations with the coefficients {\small $C_{2n}$}'s as the unknown variables:
\begin{equation*}
\small
\boldsymbol{\Delta}(\mu)
\begin{bmatrix}
 \cdots,~C_{-6},~C_{-4},~C_{-2}~C_0,~C_2,~C_4,~C_6,~\cdots
 \end{bmatrix}^T
  =
 \mathbf{0},
\end{equation*}
where {\small $\mathbf{0}$} is an infinity-dimensional zero column vector and
\begin{equation}
\small
    \boldsymbol{\Delta}(\mu)
:=
    \begin{bmatrix}
\ddots  & \vdots & \vdots & \vdots & \vdots & \vdots & \vdots & \vdots & \reflectbox{$\ddots$} \\ 
\cdots &0 & \beta_{-1} & 1 &\beta_{-1} &0 &0 &0 &\cdots \\
\cdots &0 &0 & \beta_0 & 1 &\beta_0 &0 &0 &\cdots \\
\cdots  &0 &0 &0 & \beta_1 & 1 &\beta_1 &0 &\cdots \\
\reflectbox{$\ddots$}  & \vdots & \vdots & \vdots & \vdots & \vdots  & \vdots & \vdots & \ddots \\ 
 \end{bmatrix}.
\end{equation}

This set of linear equations have nontrivial solutions
for the unknown coefficients {\small $C_{2n}$}'s 
if the determinant of {\small $\boldsymbol{\Delta}(\mu)$}, denoted as {\small $ \text{det}( \boldsymbol{\Delta}(\mu) )$}, equals zero.
From \cite{Hills_det_Simplification_bateman1953higher}, we know $\small \text{det}(\boldsymbol{\Delta} (\mu)  ) =0$ can be compactly expressed as:
{\small $    2 |\boldsymbol{\Delta}(0)| \sin^2{(\tfrac{\pi \sqrt{c_0}}{2})}  = 1- \cosh ({\mu \pi}).
    \label{Eq: Characteristic exponent_Simplified}$}
Accordingly, the exact analytical expression of {\small $\mu$} is:
\begin{equation}
\small
\mu =\frac{1}{\pi} \cosh^{-1}({1-2  |\boldsymbol{\Delta}(0)|  \sin^2{(\tfrac{\pi \sqrt{c_0}}{2})}}).
\label{Eq: mu}
\end{equation}

\noindent \textbf{Remark 2 \YanMod{(Offline computation of parameters {\small $\mu$} and {\small $\boldsymbol{\Delta}(\mu)$)}:}}
\YanMod{Recall that {\small $c_0 := -\frac{4g}{\omega^2 z_0}$ and $c_1:=\frac{2A}{z_0}$}.
Thus, the values of {\small$c_0$} and {\small$c_1$} are known
if the user-specified CoM height {\small$z_0$} is known and if the surface motion frequency {\small$\omega$} and magnitude {\small$A$} are measured or estimated in real-time (e.g., by a surface motion monitoring system~\cite{Ship_Motion_Monitoring_System}).
With known {\small$c_0$} and {\small$c_1$}, the values of {\small $\beta_n(0)$} (for all {\small $n \in \mathbb{Z}^+$}) and {\small $|\boldsymbol{\Delta}(0)|$} are known.
Then,}
we can pre-compute {\small $\mu$} using~\eqref{Eq: mu},
which could then be used to compute the analytical solution during online planning.

\subsubsection{Truncating infinite series}
\label{Sec: solution trunc.}

\YanMod{The exact solution in \eqref{Eq-Assumed_GenSol_of_MathieuEqn} is the sum of two infinite series that absolutely and uniformly converge for any {\small $0<\tau < \infty$}~\cite{dougall1915solution}. 
Thus, the solution is convergent and can be approximated as a sum of finite terms.}

With {\small $N$} terms kept,
the approximate solution is given by:
\begin{equation}
\small
    \hat{x}_{sc}(\tau) = \alpha_1 e^{\mu \tau}
    {\sum_{n=-N}^{N}} C_{2n} e^{i2n\tau} +\alpha_2 e^{-\mu \tau}\sum_{n=-N}^{N}C_{2n} e^{-i2n\tau}.
    \label{Eq-Assumed_GenSol_of_MathieuEqn_approx}
\end{equation}

\YanMod{To simultaneously ensure sufficient accuracy and efficiency for the solution computation, we can determine the number of terms kept, {\small $N$}, offline for the considered range of DRS motion parameters and the user-specified solution tolerance.
Specifically, 
we can numerically compute the minimum number of terms kept that results in a series truncation error less than the tolerance for the given DRS parameter range, and then we can set {\small $N$} as that number.}

\subsubsection{Computing {coefficients} {\small $C_{2n}$}, {\small $\alpha_1$}, and {\small $\alpha_2$}}

With {\small $\mu$} computed, we can determine the value of the coefficients {\small $C_{2n}$} ({\small $n \in \{0,1,...,N\}$}) recursively from \eqref{Eq: recurrence relation} by setting {\small $C_{2N}=0$} and {\small $C_0=A$}~\cite{BookB_ME_werth2005charged}.
The values of {\small $\alpha_1$ and $\alpha_2$} can be obtained based on the given initial conditions {\small $\hat{x}_{sc}(0)$} and {\small $\dot{\hat{x}}_{sc}(0)$}. 
The computation details are given in \AmirMod{Appendices \ref{Appendix_C2n} and \ref{Appendix_alpha}.}

\vspace{-0.1 in}
\subsection{Stability Analysis}
\label{subsec: Stability}
\vspace{-0.05 in}

By the Floquet theory~\cite{floquet1883equations}, the DRS-LIP in~\eqref{Eq-Hills_vs_motion} is called ``stable'' if all its solutions are bounded for all {\small $t>0$}, and is ``unstable'' if an unbounded solution exists for {\small $t>0$}.
The stability properties of the DRS-LIP can be determined with the characteristic exponents {\small $\mu$}.
Since the DRS-LIP is a linear, second-order ordinary differential equation, it has two characteristic exponents, denoted as {\small $\mu_1$} and {\small $\mu_2$}.
Let {\small $\text{Re}(\mu_1)$} and {\small $\text{Re}(\mu_2)$} respectively denote the real parts of {\small $\mu_1$} and {\small $\mu_2$}.
Suppose that {\small $\text{Re}(\mu_1) \leq \text{Re}(\mu_2)$}.
By the Floquet theory, the model is stable if and only if {\small $\text{Re}(\mu_1),\text{Re}(\mu_2) <0$}.
Investigation of the DRS-LIP model stability is presented in Sec.~\ref{sec: results}.

\vspace{-0.1 in}
\section{DRS-LIP BASED HIERARCHICAL PLANNING}
\vspace{-0.05 in}
\label{sec: planning}

To demonstrate the practical uses of the DRS-LIP model and its analytical solution, this section presents a hierarchical walking pattern generator that exploits them to enable efficient and feasible planning of quadrupedal DRS walking.

\begin{figure}[t]
    \centering    \includegraphics[width=1\linewidth]{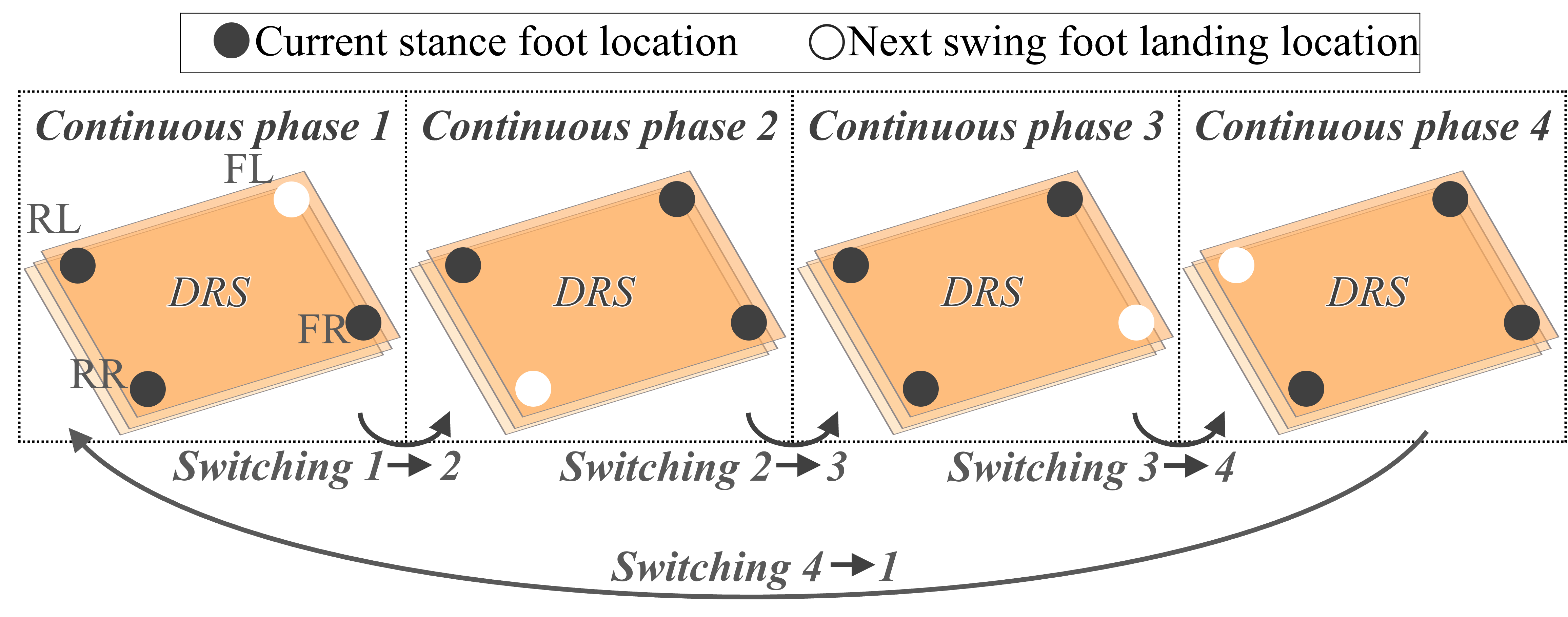}
    \vspace{-0.3 in}
    \caption{\YanMod{A complete quadrupedal walking cycle, with the four feet marked as Front Left (FL), Front Right (FR), Rear Left (RL), and Rear Right (RR).}}
    \label{Fig:Walking_phases}
    \vspace{-0.2 in}
\end{figure}

The planner is designed for quadrupedal walking~\cite{iqbal2020provably,mastalli2020motion} whose gait cycle comprises four continuous foot-swinging phases and four discrete foot-landing events (see Fig.~\ref{Fig:Walking_phases}).
This planner also assumes a known DRS motion, which is realistic for real-world applications as explained in Sec.~\ref{sec: model}.

\begin{figure*}[t]
    \centering
    \includegraphics[width=1\linewidth]{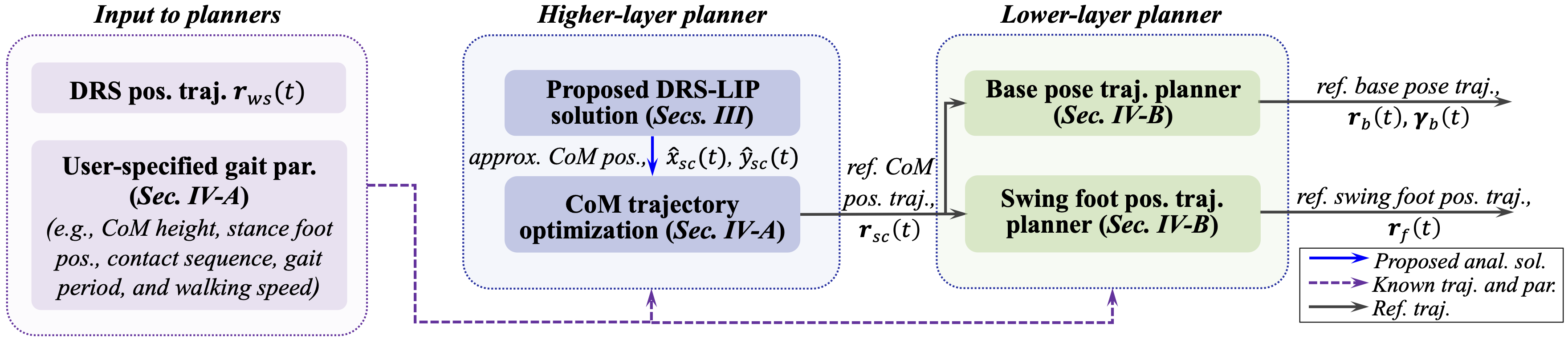}
    \vspace{-0.3 in}
    \caption{\YanMod{Overview of the proposed hierarchical walking pattern generator. The higher layer exploits the proposed analytical solution of the DRS-LIP model to ensure efficient and physically feasible planning of the desired CoM position trajectories $\mathbf{r}_{sc}(t)$. The lower layer converts the reference CoM trajectories $\mathbf{r}_{sc}(t)$ into full-body reference motions ($\mathbf{r}_{b}(t)$, $\boldsymbol{\gamma}_b(t)$, and $\mathbf{r}_f(t)$) through trajectory interpolation.}}
    \label{Fig:framework}
    \vspace{-0.2 in}
\end{figure*}

The planner has two layers (see Fig.~\ref{Fig:framework}).
The higher layer produces kinematically and dynamically feasible CoM position trajectories for the DRS-LIP model of a legged robot by incorporating necessary feasibility constraints.
The lower layer uses trajectory interpolation to efficiently translate the CoM trajectories
into the desired motion for all degrees of freedom of the full-order robot model.

\vspace{-0.12 in}
\subsection{Higher-Layer CoM Trajectory Planner}
\vspace{-0.05 in}

The higher-layer planner uses the DRS-LIP as a basis to efficiently generate feasible reference trajectories of the CoM position \YanMod{{\small $\mathbf{r}_{sc}(t)$}} through nonlinear optimization.

\subsubsection{User-defined gait parameters}
The input to the higher-layer planner is the user-defined gait parameters (which specify the desired gait features) and the known DRS motion (which is vertical and sinusoidal).
The gait parameters commonly include:
(i) average walking velocity (i.e., horizontal CoM velocity),
(ii) foot contact sequence \YanMod{(see Fig. \ref{Fig:Walking_phases})},
(iii) stance foot positions,
(iv) constant CoM height {\small $z_0$} above the surface (for respecting assumption (A2)),
and
(v) gait period.
The values of parameters (i)-(iv) are typically set to help ensure a kinematically feasible gait.
The value of the parameter (v) is selected such that the quotient of the DRS' motion period and the desired gait period is an integer (i.e., the desired CoM motion complies with the DRS motion).

\subsubsection{Optimization variables}
We choose the optimization variables {\footnotesize $\boldsymbol{\alpha}$} of the planner as the initial CoM position
\YanMod{{\small ($x_{sc}$, $y_{sc}$)}}
and velocity
\YanMod{{\small($\dot{x}_{sc}$, $\dot{y}_{sc}$)}}
within each continuous phase.
The rationale for this choice is that the DRS-LIP model parameters and these variables completely determine the horizontal CoM position trajectories.
The vertical CoM position {\small\YanMod{$z_{sc}$}} is not included as an optimization variable because it can be readily obtained from the user-defined CoM height {\small $z_0$}.

\subsubsection{Constraints}
We choose to design the constraints to help enforce gait feasibility and to respect the desired gait features specified by the user-defined parameters. 
Note that these constraints are formed based on the proposed analytical approximate solution {\small $\hat{x}_{sc}(\tau)$}.
The equality constraints include:
(i) continuity of the CoM trajectories at the foot-landing events
and
(ii) the user-specified walking velocity.
The inequality constraints are:
(i) friction cone constraint for avoiding foot slipping,
(ii) confinement of CoM trajectories within the polygon of support
for approximately respecting the CoP constraint, and
(iii) upper and lower bounds on {\footnotesize $\boldsymbol{\alpha}$}.

To meet the constraints, {\footnotesize $\boldsymbol{\alpha}$} is \YanMod{obtained by solving the following optimization problem}:
\begin{equation}
\small
\begin{aligned}
\small
\min_{\boldsymbol{\alpha}} \quad &  \small h(\boldsymbol{\alpha})
\\
\small
\textrm{subject to} \quad &  \mathbf{f}_{eq}(\boldsymbol{\alpha}) =\mathbf{0},
~
\mathbf{g}_{ineq}(\boldsymbol{\alpha}) \leq \mathbf{0},
  \label{Optimization_P1}
\end{aligned}
\end{equation}
where {\small $h(\footnotesize \boldsymbol{\alpha})$} is a scalar cost function (e.g., energy cost of transport), and the vector functions {\small $\mathbf{f}_{eq}$} and {\small $\mathbf{g}_{ineq}$} are the sets of all aforementioned equality and inequality constraints, respectively.
\YanMod{The expressions of {\small $\mathbf{f}_{eq}$} and {\small $\mathbf{g}_{ineq}$} are omitted for space consideration.}

\vspace{-0.12 in}
\subsection{Lower-Layer Full-Body Trajectory Planner}
\vspace{-0.05 in}

The lower-layer planning is essentially trajectory interpolation that translates the reference CoM trajectory {\small $\mathbf{r}_{sc}(t)$} (supplied by the higher-layer planner) into the full-order trajectories of a quadrupedal robot.
To impose a steady trunk/base pose and to avoid swing foot scuffing on the surface, we choose these full-order trajectories to be the absolute base pose (position {\small $\mathbf{r}_b$} and orientation {\small $\boldsymbol{\gamma}_b$}) and the swing foot position {\small $\mathbf{r}_f$} relative to the base.

The input to the lower-layer planner (see Fig. \ref{Fig:framework}) are: the known DRS motion that is vertical and sinusoidal, the  CoM position trajectories provided by the higher-layer planner, and user-defined parameters (e.g., CoM height, stance foot locations, and maximum swing foot height).

\subsubsection{Base \YanMod{pose} trajectories}
The CoM of the robot is approximated as the base (i.e., the geometric center of the trunk) because a quadruped's trunk typically has a symmetric mass distribution and is substantially heavier than the legs.
\YanMod{Thus, we set the desired base position trajectories {\small $\mathbf{r}_b(t)$} to be equal to the desired CoM position trajectories  {\small $\mathbf{r}_{sc}(t)$}}.
\YanMod{As real-world locomotion tasks are typically encoded by a robot's absolute global/base position, we choose to transform these relative position trajectories into the absolute ones.}

\YanMod{With the known DRS position {\small $\mathbf{r}_{ws}(t)$} at the support point {\small $S$}, 
we obtain the absolute base position trajectories {\small $\mathbf{r}_b(t)$} as:}
\begin{equation}
\small
\YanMod{\mathbf{r}_b(t)=\mathbf{r}_{sc}(t)+\mathbf{r}_{ws}(t).}
\label{Eq:base-xy trajectory}
\end{equation}
To avoid overly stretched leg joints for ensuring kinematic feasibility, the desired base orientation trajectories {\footnotesize $\boldsymbol{\gamma}_b(t)$} are designed to comply with the DRS orientation.

\subsubsection{Swing foot \YanMod{position} trajectories}
The \YanMod{desired} swing foot trajectories {\small $\mathbf{r}_{f}(t)$} \YanMod({relative to the support point {\small $S$})} are designed to agree with the user-defined stance foot locations and to respect the kinematic limits of the robot's leg joints.
Specifically, we obtain the desired swing foot trajectory during a continuous phase by using B\'ezier polynomials~\cite{iqbal2020provably} to connect the adjacent desired stance foot positions.

\YanMod{Let {\small $s$} denote the scalar normalized phase variable that represents how far a walking step has progressed.
Let {\small $\mathbf{r}_{f,i}$} and {\small $\mathbf{r}_{f,e}$} respectively denote the desired swing foot locations at the initial and end instants of a continuous phase.
We assign the values of {\small $\mathbf{r}_{f,i}$} and {\small $\mathbf{r}_{f,e}$} to match the user-defined stance foot locations for the given continuous phase.}

\YanMod{Then, we use the following B\'ezier curve to express the desired swing foot position {\small $\mathbf{r}_{f}$} within the given phase:
\begin{align}
\small
    \mathbf{r}_{f}(s) = \mathbf{r}_{f,i} + \mathbf{P}(s)(\mathbf{r}_{f,e}-\mathbf{r}_{f,i}),
    \label{Eq: SF_trajectory}
\end{align}
where {\small $\mathbf{P}(s)$} is a {\small $3 \times 3$} diagonal matrix function with each diagonal term an {\small $n^{th}$}-order B\'ezier polynomial interpolation.}

\begin{figure}[t]
    \centering
    \includegraphics[width=0.85\linewidth]{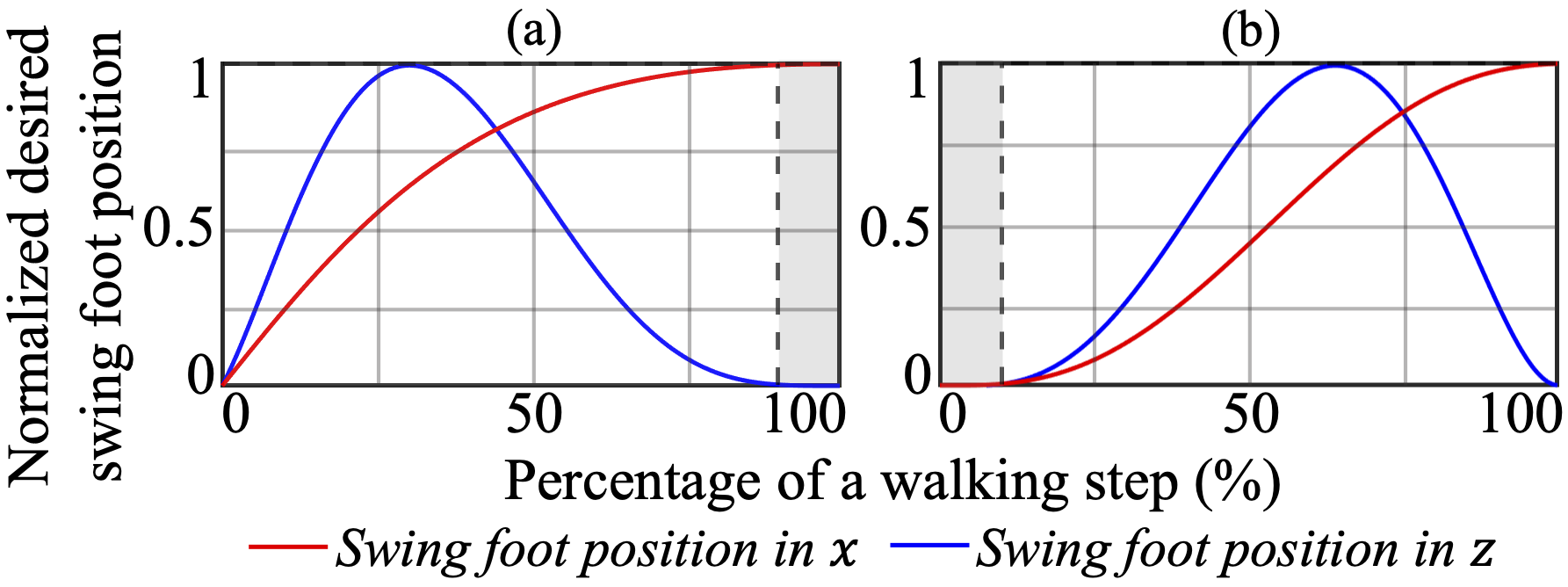}
   \vspace{-0.15 in}
    \caption{\AmirMod{Normalized swing foot position trajectories in $x$- and $z$-directions during (a) Continuous Phases 1 and 3 and (b) Continuous Phases 2 and 4.
    The grey background highlights the transitional four-leg-in-support phase.}}
    \label{Fig:SwingFootXandZ}
   \vspace{-0.25 in}
\end{figure}

\YanMod{For walking along a straight line, we can design the lateral swing foot position as constant for simplicity.
We design the forward and vertical trajectories to have a relatively fast initial velocity within Continuous Phases 1 and 3, as illustrated in Fig.~\ref{Fig:SwingFootXandZ}~(a).
This relatively fast initial velocity helps ensure the robot's full body has sufficient momentum to leave the previous support polygon and enter the planned current polygon, thus indirectly meeting the CoP constraints under the user-specified contact sequence.}

Also, as inspired by previous quadrupedal robot planning~\cite{mastalli2020motion},
a brief four-leg-in-support phase is inserted upon a foot-landing event when the two consecutive polygons of support only share a common edge (i.e., ``{\it Switching {\small $1\rightarrow 2$}}'' and ``{\it Switching {\small $3\rightarrow 4$}}'' in Fig~\ref{Fig:Walking_phases}), so as to ensure smooth and feasible transitions during these events.
\YanMod{This transitional phase is highlighted with a grey background in Fig.~\ref{Fig:SwingFootXandZ}.
Thanks to this transitional phase, the initial forward and vertical swing foot velocities within Phases 2 and 4 do not need to be as fast as Phases 1 and 3 (see Fig.~\ref{Fig:SwingFootXandZ} (b)).}

\noindent \textbf{Remark 3 (\YanMod{Effects of model accuracy on planning feasibility}):}
The dynamic feasibility of the planned trajectories partly depends on the closeness between the DRS-LIP and the actual robot dynamics.
The DRS-LIP is a relatively faithful representation of an actual DRS-robot system when the robot and DRS behaviors meet the assumptions (A1)-(A3) underlying the proposed model and its solution.
Indeed, assumption (A3) holds when the known surface motion is vertical and sinusoidal,
and the planner explicitly imposes assumption (A2).
\YanMod{Moreover, as the planner enforces the desired base orientation to comply with the surface orientation for kinematic feasibility, the planned motion will reasonably respect assumption (A1) for surfaces that translate without rotary motions.
Even for real-world DRSes that rotate (e.g., vessels), the rate of the robot's centroidal angular momentum will be negligible under the typical angular movement range of those DRSes~\cite{ShipMotion_tannuri2003estimating}, thus still respecting assumption (A1).}

\vspace{-0.05 in}
\section{SIMULATION AND EXPERIMENT VALIDATION}
\label{sec: results}

This section presents the simulation and experiment results that validate the proposed DRS-LIP model, analytical solution, and hierarchical planner.

\vspace{-0.1 in}
\subsection{Solution Validation}
\vspace{-0.05 in}

\subsubsection{{Validation of solution accuracy and efficiency}} 

The accuracy and computational efficiency of the proposed analytical approximate solution in \eqref{Eq-Assumed_GenSol_of_MathieuEqn_approx} is assessed through comparison with the highly accurate numerical solution.
For fairness of comparison, both solutions are computed in MATLAB on {\small $t \in [0, ~0.5]$} sec. 
The approximate solution has ten terms kept (i.e., {\small $N=10$}) for a reasonable trade-off between accuracy and computational efficiency \YanMod{(see Sec. \ref{Sec: solution trunc.})}.
The comparative numerical solution is computed using MATLAB's ODE45 solver with an error tolerance of {\small $10^{-9}$} and at a time interval of {\small $0.5$} ms.

To validate the proposed solution under different initial conditions, {\small 1000} sets of initial conditions are randomly chosen within \YanMod{a common movement range of quadrupedal walking~\cite{mastalli2020motion}:} {\small $|x_{sc}(0)|<0.2$} m and {\small $|\dot{x}_{sc}(0)|<0.2$} m/s.
The DRS-LIP model parameters are chosen to be within realistic ranges of DRS motions~\cite{Vessels_Vertical_Periodic_1954,ShipMotion_tannuri2003estimating} and quadrupedal robot dimensions~\cite{iqbal2020provably}: {\small $A =7$} cm, {\small $\omega = \pi$} rad/s, and {\small $z_0 =42$} cm.

\begin{figure}[t]
    \centering
    \includegraphics[width= 0.95\linewidth]{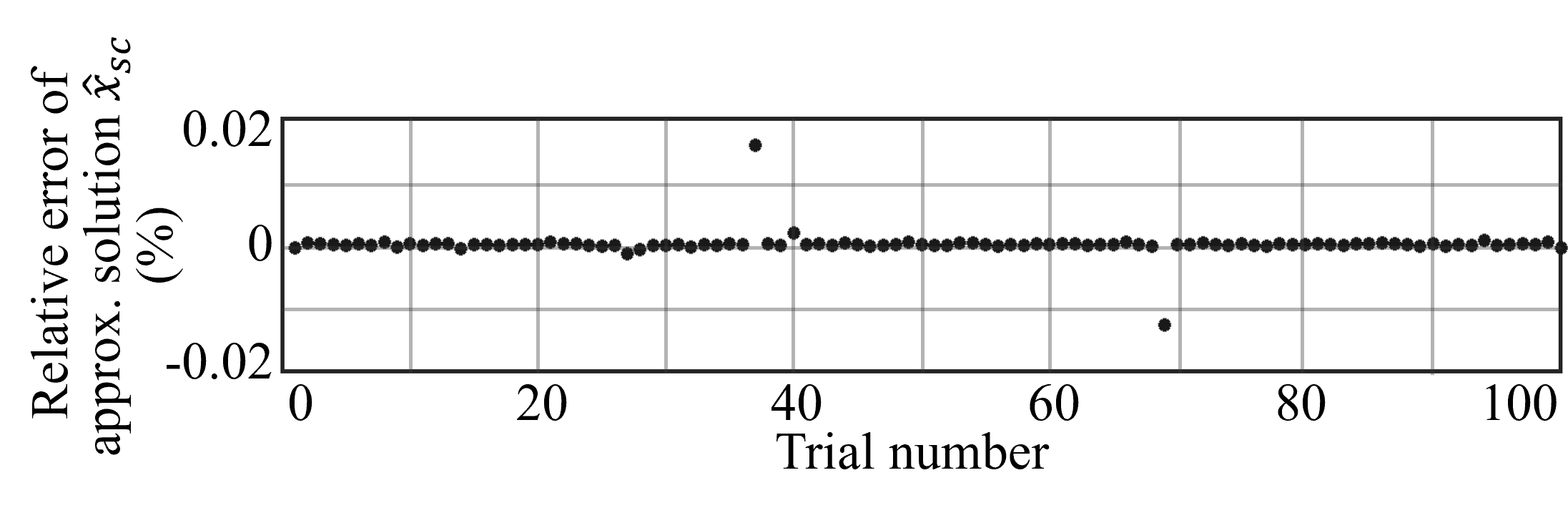}
    \vspace{-0.2in}
    \caption{\YanMod{Mean percentage error of the proposed analytical approximate solution compared with the high-accuracy numerical solution under model parameters {$A =7$} cm, {$\omega = \pi$} rad/s, and {$z_0 =42$} cm for 100 random initial conditions satisfying {$|x_{sc}(0)|<0.2$} m and {$|\dot{x}_{sc}(0)|<0.2$} m/s.}}
    \label{Fig-SolAccuracy}
    \vspace{-0.15 in}
\end{figure}

Figure~\ref{Fig-SolAccuracy} shows the accuracy of the approximate analytical solution (with ten terms kept) compared with the numerical solution for {\small 100} out of the {\small 1000} trials.
Within those {\small 100} trials, the maximum value of the absolute percentage error is lower than {\small 0.02$\%$} in magnitude, indicating the reasonable accuracy of the proposed approximate solution.
For all {\small $1000$} trials, the absolute percentage error, measured by mean $\pm$ one standard deviation (SD), is {\small $(0.0012 \pm 0.005)\%$}.

\textcolor{black}{Table \ref{table:Comparision_time_Sol} displays the comparison of the average computational time cost (measured by mean$\pm$SD) for the aforementioned {\small 1000} trials.
The approximate analytical solution is about {\small $15$} times faster to compute than the numerical one.}
      
\begin{table}[t!]
\centering
\textcolor{black}{
\caption{Average computation time of analytical and numerical solutions for 1000 trials in MATLAB (mean $\pm$ SD)}
\label{table:Comparision_time_Sol}}
\vspace{-0.12 in}
\textcolor{black}{
\begin{tabular}{c| c c } 
 \hline
  \hline
 \small  Solution method & \small  Computation time (ms) 
\\[0.5ex]
 \hline\hline 
  \small Numerical  & {\small  $2.61\pm 0.43$}  \\[0ex]  
  \small Analytical \YanMod{(proposed)} & {\small  $0.16\pm 0.02$}  \\ [0ex] 
 \hline
\end{tabular}}
\vspace{-0.1 in}
\end{table}

\begin{figure}[t]
    \centering
    \includegraphics[width= 0.95\linewidth]{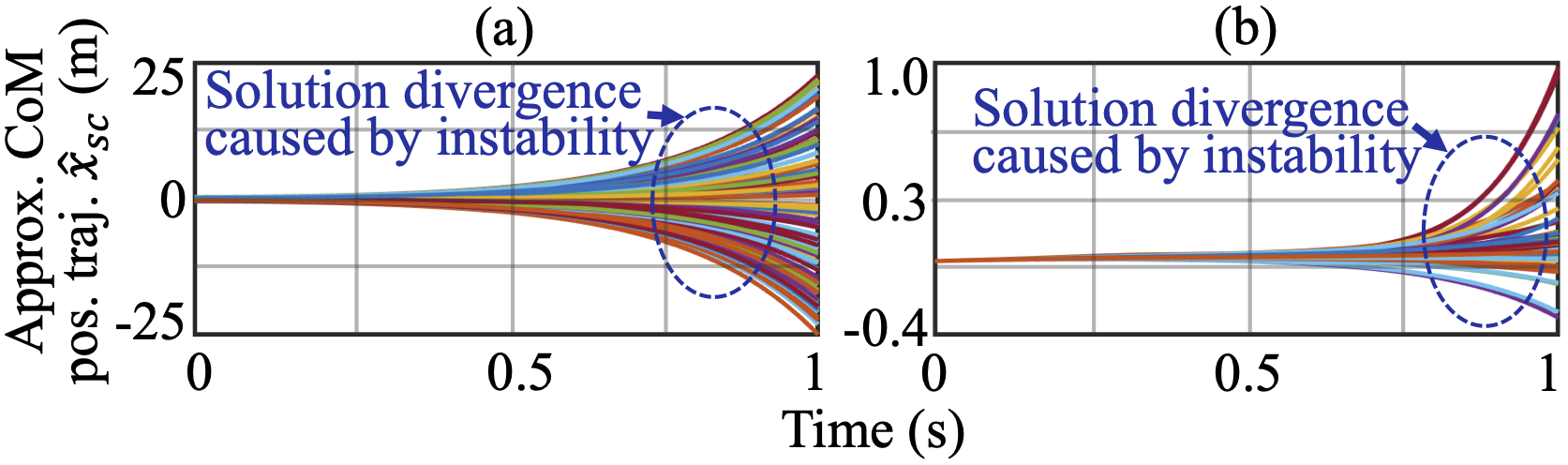}
    \vspace{-0.15 in}
    \caption{\YanMod{Unbounded time evolution of solution $\hat{x}_{sc}(t)$ of the DRS-LIP model under:
    (a) the same model parameters ({$A =7$} cm, {$\omega = \pi$} rad/s, and {$z_0 =42$} cm) but 100 different initial conditions satisfying $|x_{sc}(0)|<0.4$ m and $|\dot{x}_{sc}(0)|<0.4$ m/s
    and
    (b) different parameters ($0 <\omega\leq 2\pi$ rad/s, $0<A\leq 100$ cm, and $30\leq z_0 \leq 55$ cm) but the same initial condition ($x_{sc}(0)=0.02$ m and $\dot{x}_{sc}(0)=0.1$ m/s).}}
    \vspace{-0.2 in}
    \label{Fig-Sol_stability}
\end{figure}

\subsubsection{{Validation of stability property}}

For typical ship motions in regular sea waves~\cite{ShipMotion_tannuri2003estimating}, the parameters of the DRS-LIP model in \eqref{Eq-transformed_MathieuEqn} take values within: {\small $A \leq 100$} cm and {\small $\omega \leq 2\pi$} rad/s. 
Also, the kinematically feasible CoM height {\small $z_0$} of a typical quadrupedal robot (e.g., Unitree's Laikago) is within {\small $[0.3, 0.55]$} m.
Under these parameter ranges, we use~\eqref{Eq: mu} to numerically compute the characteristic exponents and obtain that {\small $\text{Re}(\mu_2)>0$} and {\small $\text{Re}(\mu_1) <0$}.
Thus, by the Floquet theory~\cite{floquet1883equations}, the DRS-LIP is unstable (i.e., an unbounded solution exists) under the considered operating condition.

\YanMod{To illustrate this physical insight, Fig.~\ref{Fig-Sol_stability} presents the corresponding approximate analytical solutions.}
Subplot (a) displays the approximate solutions under different initial conditions ({\small $|x_{sc}(0)|<0.4$} m and {\small $|\dot{x}_{sc}(0)|<0.4$} m/s) and DRS-LIP parameters ({\small $\omega=\pi$}  rad/s, $A=7$ cm, and {\small $z_0=42$} cm).
Subplot (b) shows the solutions under the same initial condition ({\small $x_{sc}(0) =0.02$} m and {\small $\dot{x}_{sc}(0) =0.1$} m/s) but different model parameters ({\small $0 <\omega\leq 2\pi$} rad/s, {\small $0<A\leq 100$} cm, and {\small $30\leq z_0 \leq 55$} cm).
In all cases except for the trivial initial condition {\small $x_{sc}(0),~\dot{x}_{sc}(0) = 0$}, the solutions grow towards infinity as time increases, confirming that the DRS-LIP is unstable under the considered operating condition.

\noindent \textbf{\YanMod{Remark 4 (Effects of DRS-LIP model stability on robot walking stability}):}
\YanMod{
Despite the instability of the DRS-LIP model during continuous phases, the model is useful for the planning and control of a full-order robot to ensure robot walking stability.
This is essentially because as long as the desired CoM motion is feasible during continuous phases, there exists a wide class of nonlinear control approaches (e.g., our prior input-output linearizing controller~\cite{iqbal2020provably,gu2022global}) that can provably guarantee the walking stability for the overall hybrid full-order robot model.
In this study, we implement such a controller to indirectly validate the feasibility of the proposed planner (see Sec.~\ref{Validation_planner_feasibility}).}

\vspace{-0.1 in}
\subsection{Planner Validation}
\vspace{-0.05 in}

The efficiency and feasibility of the proposed planner are validated through PyBullet simulations and experiments.

\subsubsection{{Simulation and experimental setup}}

\begin{figure}[t]
    \centering
    \includegraphics[width= 0.8\linewidth]{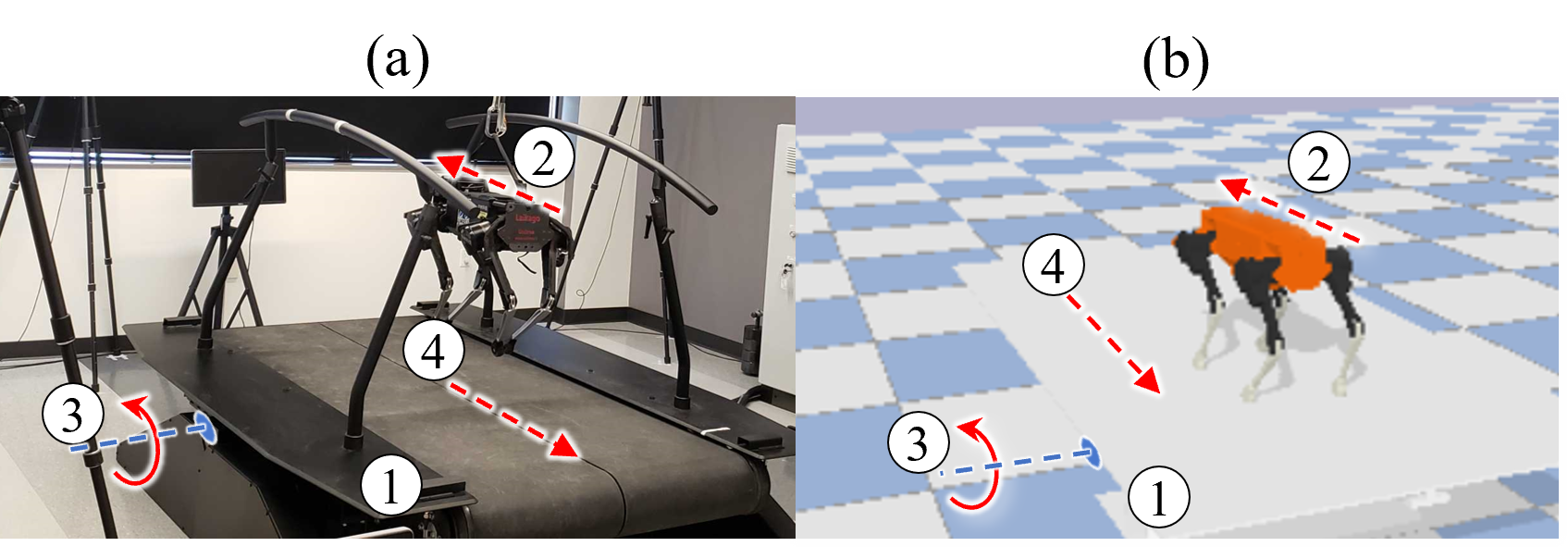}
        \vspace{-0.15 in}
    \caption{Setup of (a) experiments and (b) PyBullet simulations for testing the planner effectiveness \YanMod{using a Motek treadmill (\textcircled{1}) and a Laikago quadruped (\textcircled{2})}.
    The treadmill has a split belt (\textcircled{4}) that moves at a constant speed while the treadmill rocks about the horizontal axis (\textcircled{3}).}
    \label{Fig:Simulated_exp_setup}
    \vspace{-0.2 in}
\end{figure}

The validation of the planner utilizes a Laikago quadruped (see Fig.~\ref{Fig:Simulated_exp_setup}) developed by Unitree Robotics.
The dimension of the robot is {\small 55} cm $\times$ {\small 35} cm $\times$ {\small 60} cm.
The robot's total mass is {\small 25} kg.
It has twelve independently actuated joints.
Each leg weighs {\small 2.9} kg and has three motors located close to the trunk. 
The torque limits of the three hip-roll, hip-pitch, and knee-pitch motors are {\small 20} Nm, {\small 55} Nm, and {\small 55} Nm, respectively.

{\small
\begin{table}[t]
    \centering
    \caption{User-defined gait parameters in motion planning.}
    \label{table:Planer parameters}
    \vspace{-0.1 in}
    \begin{tabularx}{\columnwidth}{c|X X}
     \hline
      \hline
  \small   Gait parameter     &\small   (G1)  &\small  \AmirMod{(G2)}
    \\
     \hline
    \hline
     \small Friction coefficient   &\small 0.5  &\small 0.5
     \\
    \small  Robot's base height $z_0$ (cm)
       &\small 42    &\small 42 
     \\
    \small  Gait duration (s)  &\small  2    &\small  2 
     \\
    \small  Average walking velocity (cm/s)   &\small  5    &\small  6 
     \\
     \small Step length (cm)   &\small  10    &\small  12 \\
    \small  Max. step height (cm)    &\small  5   &\small  5
     \\
     \hline
    \end{tabularx}
    \vspace{-0.15 in}
\end{table}}

\begin{figure}[t]
    \centering
    \includegraphics[width= 0.85\linewidth]{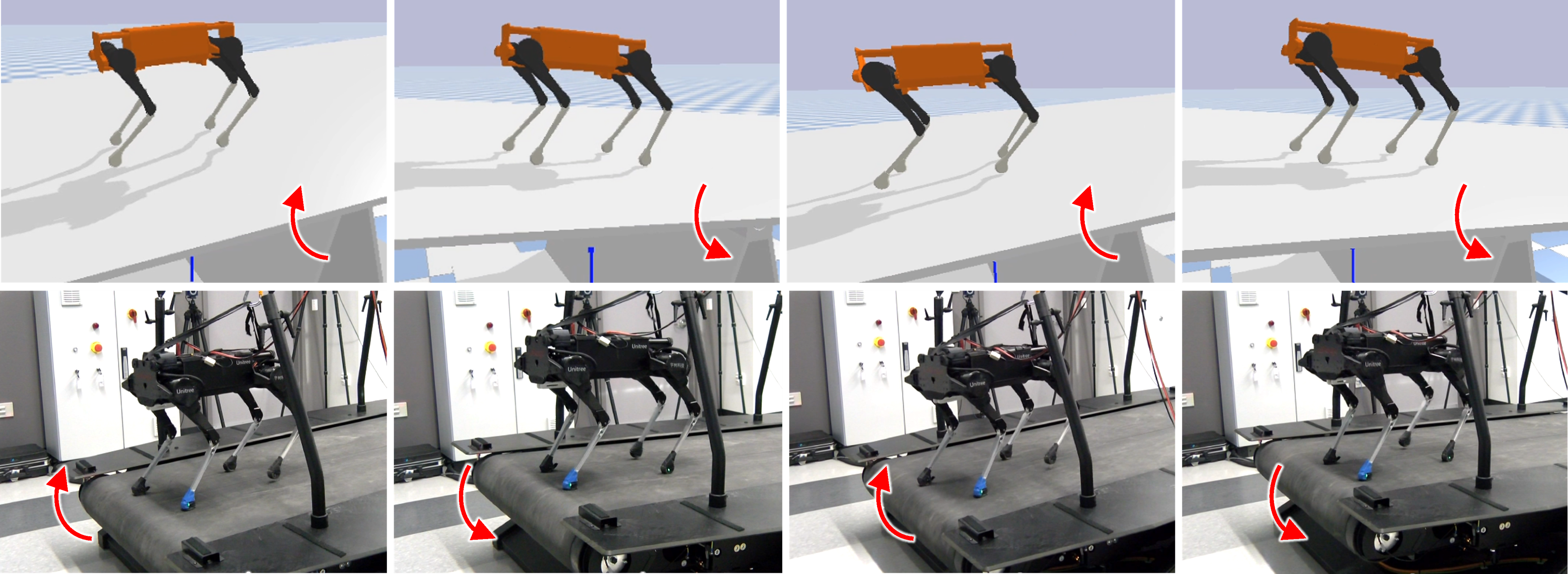}
    \vspace{-0.12 in}
    \caption{\AmirMod{Image tiles of different walking phases under gait parameters (G1) and surface motion (DRS2). The top and bottom rows respectively show PyBullet simulations and hardware experiments.}}
    \label{Fig:G1_DRS2_SnapShot}
    \vspace{-0.2 in}
\end{figure}

\noindent \textbf{DRS motion.}
Three DRS motions are tested to assess the effectiveness of the planner under different surface motions \YanMod{that emulate vessel movements in regular sea waves~\cite{Vessels_Vertical_Periodic_1954}}:
\begin{itemize}
    \item [(DRS1)]The DRS motion is vertical and sinusoidal with {\small $A=10$} cm and {\small $\omega=\pi$} rad/s.
    \item[(DRS2)] The DRS motion is a sinusoidal pitching motion with an amplitude of {\small $ 5^{\circ}$} and frequency of {\small $0.5$} Hz.
    \AmirMod{\item[(DRS3)] The DRS motion is a sinusoidal pitching motion with an amplitude of {\small $ 7^{\circ}$} and frequency of {\small $0.5$ Hz}}.
\end{itemize}
Surface motions (DRS2) and \YanMod{(DRS3)} reasonably satisfy assumption (A3) because the associated horizontal velocities of the surface are negligible due to the small pitching amplitudes.
\YanMod{Still, in the vertical direction, the surface accelerations under (DRS1)-(DRS3) are relatively significant for planner validation, with peak contact-point accelerations approximately at {\small 100} {$\text c \text m / \text s^2$}, {\small 70} {$\text c \text m / \text s^2$}, and {\small 110} {$\text c \text m / \text s^2$} in magnitude, respectively, when the robot stands about {\small $1$} m away from the treadmill's axis of pitching.
The corresponding contact-point displacements are {\small 10} cm, {\small 7} cm, and {\small 11} cm, respectively.}

\noindent \textbf{Simulated and physical DRSes.}
To validate the planner feasibility, the surface motions (DRS2) and (DRS3) are realized both in simulations and experimentally by a physical Motek M-Gait treadmill (see Fig.~\ref{Fig:Simulated_exp_setup}), and (DRS1) is implemented in PyBullet simulations alone.
The Motek treadmill can be pre-programmed to perform user-defined pitching (but not vertical) motions and belt translation.
The treadmill weighs {\small 750} kg with a dimension of {\small 2.3} m  $\times$ {\small 1.82} m $\times$ {\small 0.5} m.
A {\small 4.5} kW servo motor powers each of the treadmill's two belts.
During the \YanMod{hardware experiments}, the robot is placed approximately {\small $1$} m away from the treadmill's axis of rotation, and
the belt speed is set to be the same as the desired walking speed.
\AmirMod{Figure ~\ref{Fig:G1_DRS2_SnapShot} shows images of the Laikago robot walking on the rocking treadmill in simulations and experiments.}

\noindent \textbf{Gait parameters.}
Recall that the proposed planner takes user-defined gait parameters as its input.
To evaluate the planner under different gait parameters, two sets of parameters (G1) and (G2) are used (see Table~\ref{table:Planer parameters}).

\subsubsection{{Validation of planner efficiency}}
To validate that using the proposed analytical solution improves the planner efficiency compared with using the numerical solution, the higher-layer CoM trajectory planning problem is solved based on both solutions under the user-defined gait parameters (G1) and surface motion (DRS2).
For simplicity, the cost function {\small ${h}$} in~\eqref{Optimization_P1} is chosen as trivial.
A {\small $6^{th}$}-order B\'ezier curve is used to design the desired swing foot trajectory for allowing adequate freedom in trajectory design.
\YanMod{Also, we choose to lower the load of computing the proposed analytical solution by pre-computing its solution parameters {\small $\mu$} and {\small $C_{2n}$}, which is realistic for practical applications where the surface motion is sensed or estimated~\cite{Ship_Motion_Monitoring_System} (Remark 2).}

To demonstrate the improved efficiency under different common solvers, both MATLAB and C++ are used to solve the optimization-based planning problem in \eqref{Optimization_P1} for {\small 1000} runs with the same initial guess of {\small $\boldsymbol{\alpha}$}.
For fairness of comparison, the optimality and constraint tolerances are set as {\small $ 10^{-6}$} in all runs.
In MATLAB, fmincon is used with an interior-point solver.
For the C++ optimization, the nonlinear optimization solver of the Ipopt package \cite{wachter2006implementation_ipopt} is utilized.

For those {\small $1000$} runs, Table \ref{table:Comparision_time} shows that the mean time costs of the analytical solution based higher-layer planning is approximately {\small \AmirMod{7}} and {\small \AmirMod{4}} times shorter than the numerical solution based one in MATLAB and C++, respectively.

Table \ref{table:Comparision_time} also indicates that the higher-layer planner takes {\small  \AmirMod{{\small$8.6 \pm 2.2$}}} ms to generate the desired CoM trajectory when it is solved by C++ using the approximate analytical solution.
\AmirMod{The median time cost of those {\small $1000$} runs of computations is {\small$8.4$} ms.}
Also, solving the lower-layer planner is typically fast (e.g., MATLAB can solve it within {\small $2$} ms) since the planning is essentially trajectory interpolation.
Thus, the mean time cost for solving both higher and lower layers will be less than \AmirMod{{\small$11$}} ms.
Since such a time cost is much smaller than the typical quadrupedal walking gait period (i.e., about {\small $2$} sec~\cite{mastalli2020motion}) and real-world DRS motion periods (e.g., {\small 1-100} sec for vessels~\cite{ShipMotion_tannuri2003estimating}), the proposed planner would be adequately fast to timely regenerate the desired full-order trajectories in case of any significant changes in the DRS motion.

\begin{table}[t!]
\centering
\caption{Average time cost of 1000 runs of higher-layer planning (mean$\pm$SD) under gait (G1) and surface motion (DRS2).}
\vspace{-0.15 in}
\label{table:Comparision_time}
\begin{tabular}{c | c c } 
\hline
\hline
\small Solution method & \small MATLAB  &\small  C++ \\ 
 &\small  (fmincon) &\small  (Ipopt)\\
 \hline\hline 
\small  Numerical (ms)   & {\small  $1320.7 \pm 13.8$}  &\small  \AmirMod{$72.3 \pm 6.7$}\\[0ex]  
\small  Analytical (ms)  &\small  $269.1\pm 12.9$  &\small  \AmirMod{$8.6 \pm 2.2$}  \\ [0ex] 
 \hline
\end{tabular}
\vspace{-0.2 in}
\end{table}

\subsubsection{{Validation of planner feasibility}}
\label{Validation_planner_feasibility}
Beside efficiency, the proposed DRS-LIP and its solution also help guarantee planning feasibility.
To test the feasibility of the planned motion,
our previous input-output linearizing controller~\cite{iqbal2020provably}, which is derived based on the hybrid full-order robot model and proportional derivative (PD) control, is utilized to track the planned full-order trajectories in PyBullet simulations and hardware experiments \YanMod{(Remark 4)}.
As this controller does not explicitly ensure the feasibility of ground contact forces, the planned trajectory needs to be physically feasible in order for the controller to be effective.
Thus, if the controller is able to reliably track the planned motion and sustain walking on a DRS, then the physical feasibility of the proposed planner is confirmed.
To help ensure a reasonable tracking performance, PD gains are tuned as {\small $0.7$} and {\small $1.0$} in simulations, and {\small $5.5$} and {\small $0.15$} on hardware.

\begin{figure}[t]
    \centering
    \includegraphics[width= 0.95\linewidth]{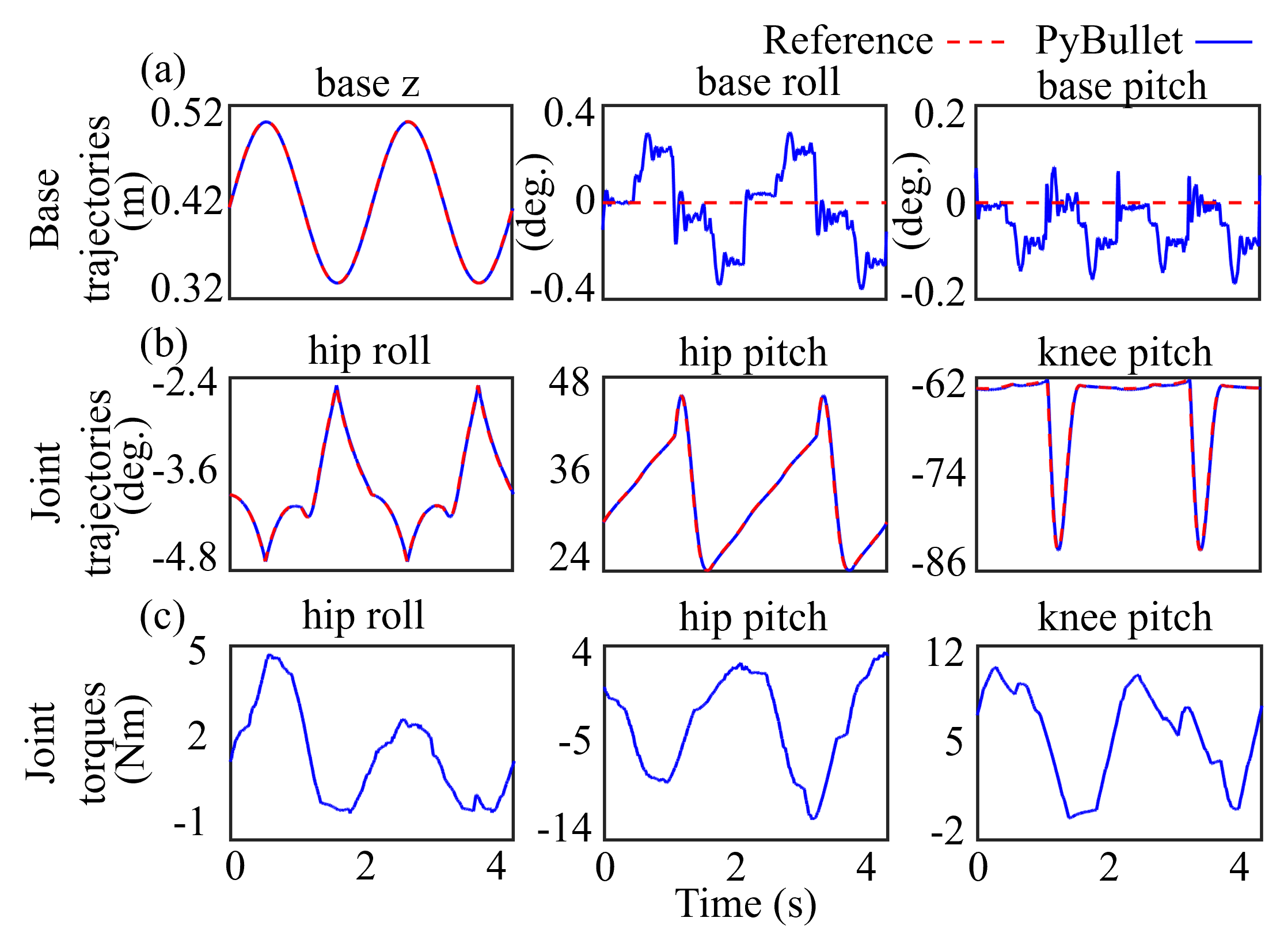}
   \vspace{-0.2 in}
    \caption{PyBullet simulation results at the robot's front-right leg under gait parameters (G1) and surface motion (DRS1).}
    \label{Fig:G1_DRS1_sim}
     \vspace{-0.2 in}
\end{figure}

\YanMod{To validate the planner feasibility
under vertical surface motions,} 
the gait parameters (G1) and the surface motion (DRS1) are tested in PyBullet simulation.
As shown in Fig. \ref{Fig:G1_DRS1_sim}, the robot sustains walking for the entire testing duration, which is over {\small 50} gait cycles.
The base and joint trajectories closely track their reference values, as shown in subplots (a) and (b). 
Also, subplot (c) indicates that the actual robot motion indeed respects the torque limits.

\begin{figure}[t]
    \centering
    \includegraphics[width= 
    0.95\linewidth]{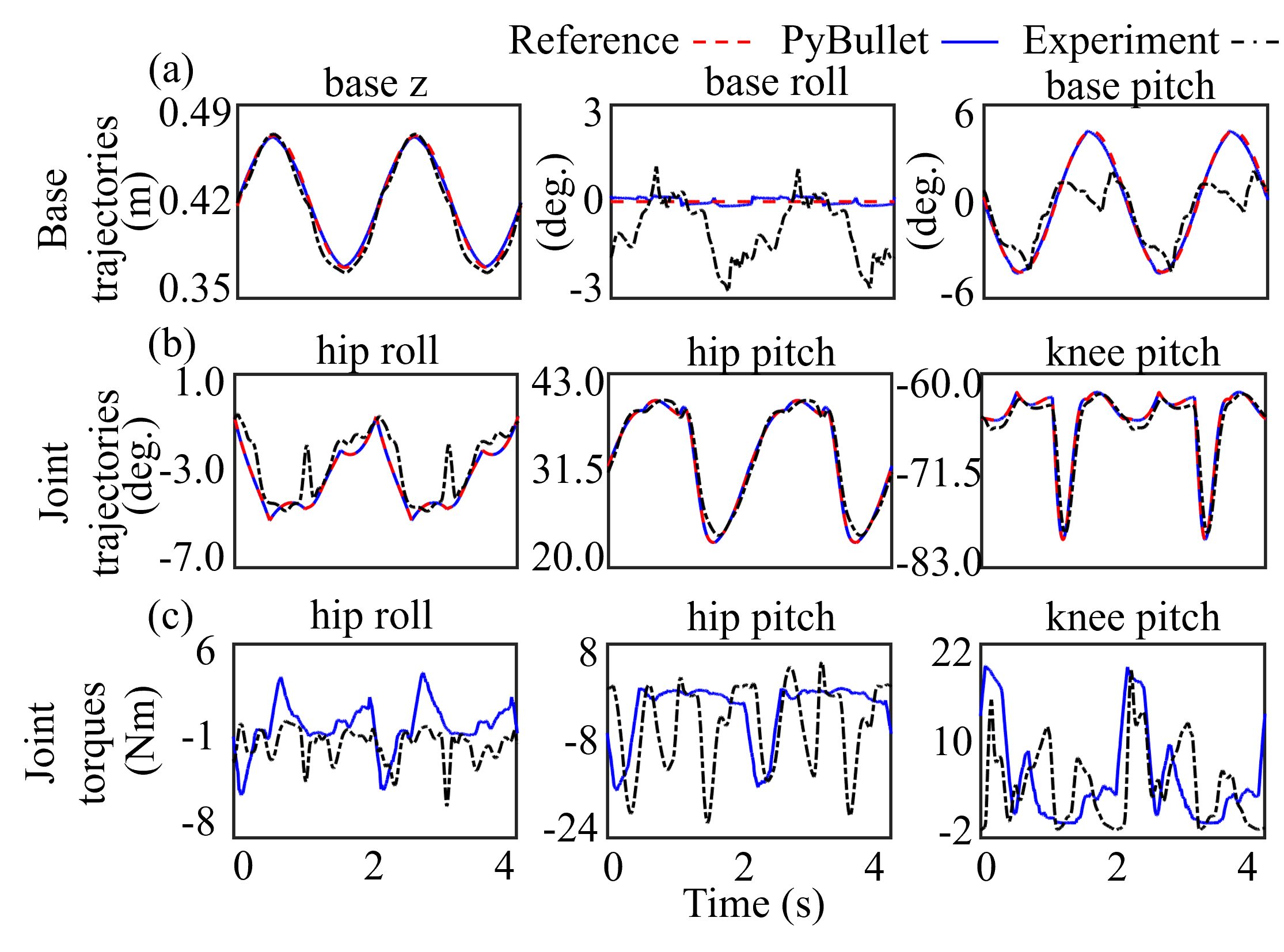}
    \vspace{-0.2 in}
    \caption{Hardware experiment and PyBullet simulation results at the robot's front-right leg under gait parameters (G1) and surface motion (DRS2).}
    \label{Fig:G1_DRS2}
    \vspace{-0.2 in}
\end{figure}

\begin{figure}[t]
    \centering
    \includegraphics[width= 0.95\linewidth]{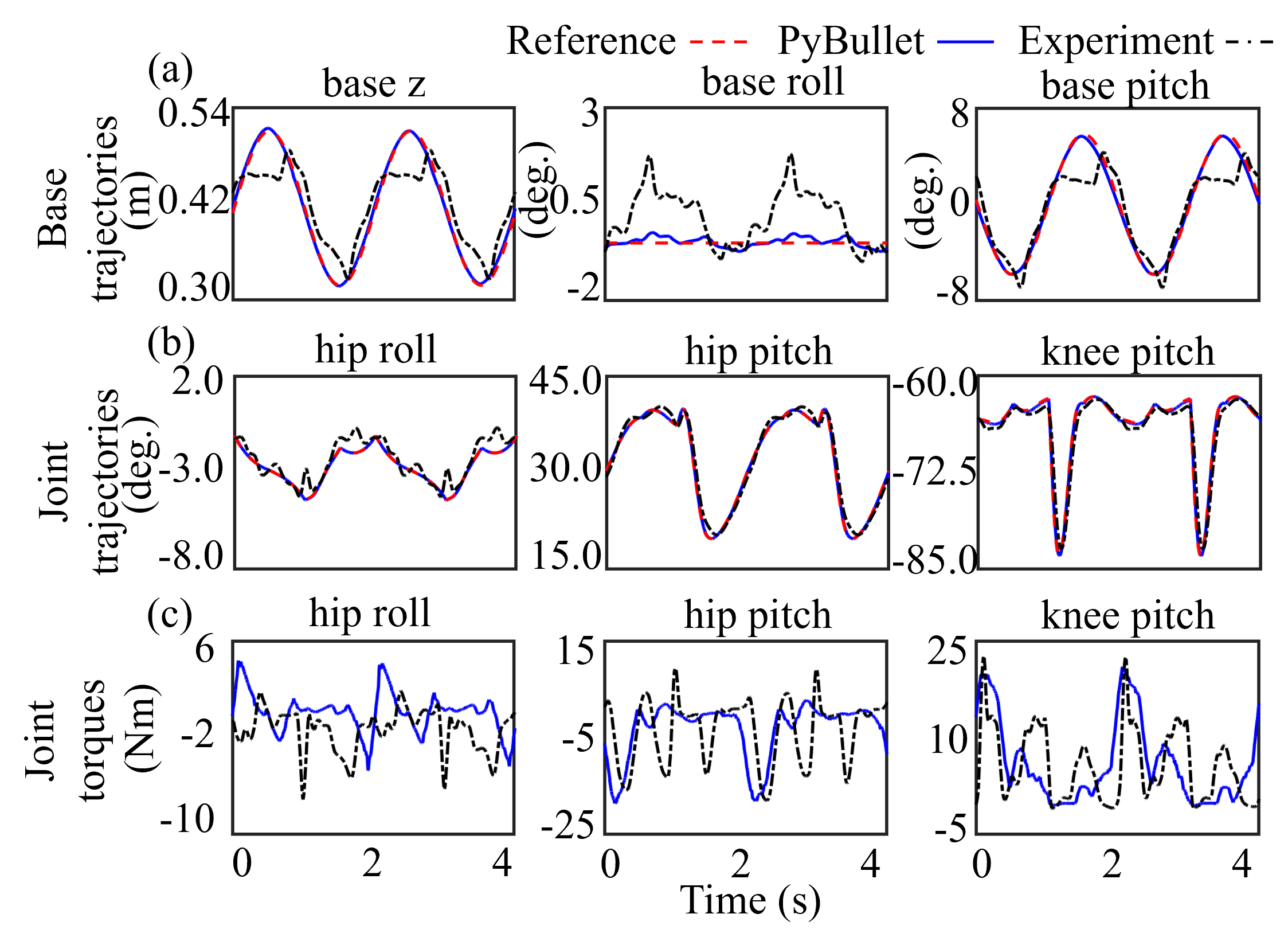}
    \vspace{-0.2 in}
    \caption{\AmirMod{Hardware experiment and PyBullet simulation results at the robot's front-right leg under gait parameters (G2) and surface motion (DRS3).}}
    \label{Fig:G2_DRS3}
    \vspace{-0.2 in}
\end{figure}

\YanMod{To further assess the planner feasibility under different gait parameters and pitching surface motions,}
the combination of (G1) and (DRS2) \AmirMod{and that of (G2) and (DRS3) are} tested in simulations and experiments, \YanMod{with the results respectively presented in Figs.~\ref{Fig:G1_DRS2} and \ref{Fig:G2_DRS3}.}
The experiment video is available at \href{https://youtu.be/u2Q_u2pR99c}{\tt \small \AmirMod{https://youtu.be/u2Q\_u2pR99c.}}
In both simulations and experiments, the robot walking is stable, as indicated by the trajectory tracking accuracy in subplots (a) and (b) as well as the experiment video.
Moreover, subplots (c) confirm that the joint torque limits are met in both simulations and experiments.
Yet, the torque profiles of the front-right leg's three joints display notable discrepancies between PyBullet and experiment results, possibly due to the differences between the simulated and actual robot dynamics as well as the different inherent meanings of their effective PD gains.
Also, the experiment video shows that the robot experiences relatively notable rebounding and slipping at contact switching events when a rear leg lands on the surface.
This violation of the planned contact sequence is directly due to the temporary loss of contact force feasibility, and could be mitigated through improved controller design as discussed in Sec.~\ref{sec: discussion}.

\vspace{-0.1 in}
\section{DISCUSSION}
\label{sec: discussion}
\vspace{-0.05 in}

This paper has introduced a reduced-order dynamic model of a legged robot that walks on a DRS, by analytically extending the classical LIP model from stationary surfaces~\cite{kajita20013d} to a DRS (e.g., a vessel).
The resulting DRS-LIP model in~\eqref{Eq-LIPM_on_DRS_simplified} is a linear, second-order differential equation, similar to the classical LIP.
However, the DRS-LIP is explicitly time-varying whereas the classical LIP is time-invariant.
This fundamental difference is due to the time-varying movement of the surface at the surface-foot contact points.
This study also \YanMod{investigates the stability} of the DRS-LIP based on the Floquet theory (see Sec.~\ref{sec: results}-A).
Similar to the classical LIP that describes stationary surface locomotion~\cite{kajita20013d}, the DRS-LIP is unstable under the usual movement range of real-world DRSes such as vessels~\cite{ShipMotion_tannuri2003estimating}.

The DRS-LIP is valid under the assumption that the actual robot's rate of whole-body angular momentum about the CoM is negligible (assumption (A1)).
To relax this assumption, the point mass of the proposed DRS-LIP could be augmented with a flywheel~\cite{pratt2006capture,zhao2017robust} to account for the nonzero rate of angular momentum.
Moreover, the DRS-LIP can be generalized from a constant CoM height (as enforced by assumption (A2)) to a varying height by integrating with the variable-height LIP for stationary surfaces~\cite{Cap_uneve_caron2019TRO}.

This study also derives the approximate analytical solution of the DRS-LIP 
for vertical, sinusoidal surface motions.
Its \YanMod{sufficient accuracy and improved} computational efficiency compared with numerical solutions are confirmed through MATLAB simulations 
(Fig.~\ref{Fig-SolAccuracy} and Table~\ref{table:Comparision_time_Sol}).
\AmirMod{Although the proposed reduced-order model in \eqref{Eq:simplified_Cap-xy} does not assume a specific form of surface motion, the proposed analytical solution is derived based on the assumption that the surface motion is vertical and sinusoidal.
Such a surface motion is typical for real-world ship motions in regular sea waves~\cite{gahlinger2000const_hv_ship,ShipMotion_tannuri2003estimating}.
To address surface motions that are vertical and nonperiodic with their time profiles pieced together by periods of different sinusoidal waves, which cover a wide range of DRS motions \cite{ShipMotion_tannuri2003estimating}, the proposed analytical solution could be extended by: a) forming the individual analytical solutions for those different periods based on the proposed solution derivation method and then b) piecing them together to form the needed overall solution.
Also, if the vertical nonperiodic surface motion is pieced together by periods of general periodic functions instead of sinusoidal waves, we could potentially use the Floquet theory \cite{floquet1883equations} to derive the analytical solution by numerically precomputing the fundamental matrix of the reduced-order model and then forming the analytical solution using the fundamental matrix.
Our future work will also tackle the modeling and planning problem for legged locomotion under general surface motions that contain horizontal movements~\cite{yuanDW2022}.}

To highlight the usefulness of the analytical results, they have been used as a basis to synthesize a hierarchical planner that efficiently produces desired, physically feasible motions for quadrupedal DRS walking. The feasibility of the planned motion is validated by using our previous tracking controller~\cite{iqbal2020provably} to command a quadrupedal robot to follow the planned motion during DRS walking.
As discussed in Sec. V-B, 
simulation and experiment results indicate the reasonable feasibility of the proposed planner under different gait parameters and surface motions (Figs. \ref{Fig:G1_DRS1_sim}-\ref{Fig:G2_DRS3}).
To mitigate the temporary violation of the planned gait sequence observed in experiments, which is partly induced by the discrepancies between the DRS-LIP and the actual robot dynamics, the planned motion could be tracked by an optimization-based controller that explicitly ensures physical feasibility.

\YanMod{The proposed planner assumes a constant average walking speed (i.e., the average horizontal speed of the robot’s base/trunk relative to the walking surface) for all walking cycles. However, in the case of variable speed walking, the robot’s desired average walking speed should vary among different walking cycles. To that end, the proposed planner should be extended from constant speed walking to variable speed walking in our future work. Such an extension would be feasible essentially because within any given walking cycle of constant or variable speed walking, the robot behaviors that are captured by the proposed reduced-order model and its analytical solution have the same mathematical expressions and because these expressions hold without any assumptions on the variability of the walking speed. A potential approach to enable such an extension is: (a) to plan desired walking cycles with different walking speeds by setting the user-defined gait parameters of both layers of the proposed planner, such as step length and duration, to vary among different walking cycles, and (b) to stitch the desired walking cycles of different average speeds for forming the overall variable speed walking motions.}

\YanMod{In the hardware experiment validation of the feasibility of the proposed planner, the desired speed of quadrupedal robot walking on a vertically moving surface is set between {\small $5$} cm/s and {\small $6$} cm/s.
This speed is, to our best knowledge, the fastest speed of quadrupedal walking on a vertically moving surface for hardware experiments~\cite{iqbal2020provably}.
To plan dynamic-surface walking with higher speed, ensuring the feasibility of the proposed reduced-order model based planner will be more challenging.
This is because, similar to other reduced-order models~\cite{kajita20013d,pratt2006capture}, the nonlinearity of the actual dynamics ignored in the LIP model will become more significant during faster walking,
resulting in a larger discrepancy between the actual robot dynamics and the LIP model and thus potentially causing physical nonfeasibility of the planned motion.
A promising solution is to augment the controller described in Sec. V, which does not explicitly guarantee the feasibility of necessary constraints (e.g., ground contact forces), with an optimization-based controller~\cite{fawcett2021robust,mastalli2020motion} that explicitly ensures the feasibility for actual walking.}

\vspace{-0.1 in}
\section{CONCLUSION}
\vspace{-0.05 in}
\label{sec: conclusion}
\YanMod{This paper has introduced a reduced-order robot dynamics model (termed as DRS-LIP), its approximate analytical solution, and a real-time motion planner for legged walking on a vertically moving DRS.
The DRS-LIP describes the essential robot dynamics associated with DRS walking, and was derived by theoretically extending the classical LIP from a stationary surface to a DRS.
Its analytical solution was obtained based on the conversion of the DRS-LIP into Mathieu's equation.
Exploiting these analytical results as a basis, a real-time planner was designed to efficiently generate feasible quadrupedal walking motions for DRS walking.
Simulation results revealed the continuous-phase stability property of the DRS-LIP and the efficiency and accuracy of the analytical solution under common real-world DRS movements.
Finally, 3-D realistic PyBullet simulations and experiments on a Laikago robot confirmed the computational efficiency and physical feasibility of the proposed planner under different gait parameters and surface motions.}

\vspace{-0.1 in}
\bibliography{ReferencesAbbrev} 
\bibliographystyle{ieeetr}

\appendices

\vspace{-0.1 in}
\AmirMod{\section{Recurrence Relationship between {\small$\mu$} and {\small$\beta_n$} for Computing the Proposed Analytical Solution}}
\label{Appendix_A}
\vspace{-0.05 in}

\YanMod{This appendix introduces the derivation of the recurrence relationship between the solution parameters {\small$\mu$} and {\small$\beta_n$} as expressed in~\eqref{Eq: recurrence relation} from Sec. III-A1.}

\YanMod{Recall that the proposed DRS-LIP model in~\eqref{Eq-Hills_vs_motion} can be rewritten as Mathieu's equation~\cite{farkas2013periodic} in \eqref{Eq-transformed_MathieuEqn};
that is,
\begin{equation*}
  \tfrac{d^2x_{sc}}{d\tau^2} +(c_0 - 2c_1\cos 2 \tau)x_{sc}  = 0.  
\end{equation*}
The solution of Mathieu's equation can be assumed as~\cite{Hills_det_Simplification_bateman1953higher}:
\begin{equation}
   \small
   {{x}_{sc}(\tau) = e^{\mu \tau}\sum_{n=-\infty}^{\infty}C_{2n} e^{i2n\tau}.}
    \label{Eq-Assumed_Sol1}
\end{equation}
Thus, we use \eqref{Eq-Assumed_Sol1} to rewrite \eqref{Eq-transformed_MathieuEqn} as:
\begin{equation*}
\small
\begin{aligned}
    \sum_{n=-\infty}^{\infty}[(\mu^2 +&(i2n)^2 +4i\mu n)C_{2n} + (c_0 -2c_1 \cos 2 \tau)C_{2n}] e^{(i2n+\mu)\tau} =0.
    \label{Eq-RR1}
\end{aligned}
\end{equation*}}
\YanMod{With {\small$\cos 2 \tau =\frac{e^{i2\tau}+e^{-i2\tau}}{2}$}, this equation becomes:
\begin{equation*}
\small
\begin{aligned}
\sum_{n=-\infty}^{\infty}[(\mu^2 &-(2n)^2 +2 (i\mu) (2n)+c_0)C_{2n} \\
   &-c_1(e^{i2\tau}+e^{-i2\tau})C_{2n}] e^{(i2n+\mu)\tau} =0,
\end{aligned}
\end{equation*}
which can be further rearranged as:
\begin{equation}
\small
\begin{aligned}
    \sum_{n=-\infty}^{\infty}[((i\mu)^2 &+(2n)^2 -2 (i\mu) (2n)-c_0)C_{2n}e^{(i2n+\mu)\tau} \\ &+c_1(C_{2n}e^{(i2(n+1)+\mu)\tau}+C_{2n}e^{(i2(n-1)+\mu)\tau}]  =0.
    \label{Eq-RR3}
\end{aligned}
\end{equation}}
\AmirMod{Since \eqref{Eq-RR3} is the sum over indices ranging from {\small$-\infty$ to $\infty$}, we can transform it into~\cite{Hills_det_Simplification_bateman1953higher}:
\begin{equation*}
\small
\begin{aligned}
    &\sum_{n=-\infty}^{\infty}[((2n-i\mu)^2-c_0)C_{2n}+ c_1C_{2(n+1)}+  c_1C_{2(n-1)}]e^{(i2n+\mu)\tau}   =0;   
    \label{Eq-RR42}\\
\end{aligned}
\end{equation*}
that is, {\small $\beta_n(\mu) C_{2(n+1)} + C_{2n} + \beta_n(\mu) C_{2(n-1)} =0$}, where {\small$\beta_n(\mu) : = \tfrac{c_1}{(2n-i\mu)^2-c_0}$}.}

\vspace{-0.1 in}
\AmirMod{\section{Computing Solution Coefficient {\small$C_{2n}$}}
\vspace{-0.1 in}
\label{Appendix_C2n}}

\YanMod{This appendix presents the computation of the solution coefficients {\small $C_{2n}$}, which is omitted in Sec.III-A3.} 

\YanMod{The recurrence relationship in \eqref{Eq: recurrence relation} indicates that the coefficient satisfies {\small$\AmirMod{|}C_{2n}\AmirMod{|}<<\AmirMod{|}C_{2(n-1)}\AmirMod{|}$} for sufficiently large index {\small$n$} (e.g., {\small$n>N$}, with {\small $N$} the number of terms kept in the approximate solution).
Thus, coefficients with sufficiently large indices can be neglected (i.e., {\small$C_{2(N+1)} \approx 0$}).}

\YanMod{With {\small$C_{2(N+1)}=0$}, solving the recurrence relation in \eqref{Eq: recurrence relation} for various indices gives:
\begin{equation*}
\small
\begin{aligned}
    \text{for}~n = N:&~~~~\beta_N C_{2(N+1)} + C_{2N} + \beta_N C_{2(N-1)} =0, \\ 
    &\Rightarrow C_{2N} = - \beta_N C_{2(N-1)},~\text{since} ~C_{2(N+1)}=0
    \\
    \text{for}~n = N-1:&~~~~\beta_{N-1} C_{2N} + C_{2(N-1)} + \beta_{N-1} C_{2(N-2)} =0, \\ 
    &\Rightarrow C_{2(N-1)} = \tfrac{- \beta_{N-1}}{1- \beta_N \beta_{N-1}} C_{2(N-2)}
    \\
    \text{for}~n = N-2:&~~~~\beta_{N-2} C_{2(N-1)} + C_{2(N-2)} + \beta_{N-2} C_{2(N-3)} =0, \\ 
    &\Rightarrow C_{2(N-2)} = {\footnotesize \tfrac{- \beta_{N-2}}{1- \tfrac{ \beta_{N-2} \beta_{N-1}}{1- \beta_{N-1}\beta_N }}}C_{2(N-3)}
    \\
    &~~~~~~~~~~~~~~\cdots
    \label{Eq-RR5}
\end{aligned}
\end{equation*}
Thus, {\small$C_{2n}$} ({\small $n \in \{0,1,...N \}$}) can be expressed as:
\begin{equation}
\small
\begin{aligned}
     C_{2n} = \cfrac{- \beta_{n} }{1- \tfrac{ \beta_{n}\beta_{n+1} }{1- \tfrac{\beta_{n+1} \beta_{n+2}}{1-\tfrac{\beta_{n+2} \beta_{(n+3)}}{1-~~\cdots}}}}C_{2(n-1)}.
    \label{Eq-Coefficient}
\end{aligned}
\end{equation}
By setting {\small$C_0 =A$} 
in \eqref{Eq-Coefficient}~\cite{phelps1965analytical}, all other coefficients can be determined using \eqref{Eq-Coefficient}.
Also, the relation in \eqref{Eq-Coefficient} can be used to find the coefficients {\small$C_{-2n}$}, by replacing index {\small $n$} with its additive inverse {\small $-n$}.
Recall that {\small$\beta_n$} is defined in Sec. III-A1 as {\small$\beta_n(\mu) = \frac{c_1}{(2n-i\mu)^2-c_0}$}.
This definition indicates that {\small$\beta_{-n}$} is the complex conjugate of {\small$\beta_n$}, and accordingly {\small$C_{-2n}$} is the complex conjugate of {\small$C_{2n}$}.}

\vspace{-0.1 in}
\AmirMod{\section{Computing Solution Coefficients {\small $\alpha_1$} and {\small $\alpha_2$} for a given initial condition}
\vspace{-0.05 in}
\label{Appendix_alpha}}

\AmirMod{From Appendix \ref{Appendix_C2n}, we know the solution coefficient {\small $C_{2n}$} is a complex number and its complex conjugate is {\small $C_{-2n}$}.
Denoting {\small $C_{2n}$} as {\small $C_{2n}= r_{2n}e^{i\theta_{2n}}$}, where {\small$r_{2n}$} and {\small$\theta_{2n}$} are real constants,
and substituting {\small $C_{2n}= r_{2n}e^{i\theta_{2n}}$}
in the approximate analytical solution \eqref{Eq-Assumed_GenSol_of_MathieuEqn_approx}, we obtain:
\begin{equation*}
\small
\begin{aligned}
    \hat{x}_{sc}(\tau) &= \alpha_1 e^{\mu \tau}\sum_{n=-N}^{N} r_{2n}e^{i\theta_{2n}} e^{i2n\tau} +\alpha_2 e^{-\mu \tau}\sum_{n=-N}^{N} r_{2n}e^{i\theta_{2n}} e^{-i2n\tau}\\
    &=\alpha_1 e^{\mu \tau}\sum_{n=1}^{N}[r_{0}+ r_{2n}( e^{i(2n\tau +\theta_{2n})}+e^{-i(2n\tau +\theta_{2n})})] 
    \\&~~~~
    +\alpha_2 e^{-\mu \tau}\sum_{n=1}^{N}[r_{0}+ r_{2n}( e^{-i(2n\tau-\theta_{2n})}+e^{i(2n\tau-\theta_{2n})})]
    \\
    &=\alpha_1 e^{\mu \tau}\sum_{n=1}^{N}[r_{0}+ 2r_{2n}\cos(2n\tau +\theta_{2n})]
    \\&~~~~
     +\alpha_2 e^{-\mu\tau}\sum_{n=1}^{N}[r_{0} + 2r_{2n}\cos(2n\tau-\theta_{2n})].
    \label{Eq-Sol_r_th_1}
\end{aligned}
\end{equation*}
Recall {\small $\tau :=\frac{\frac{\pi}{2}+\omega t}{2}$}.
Replacing {\small $\tau$} with {\small $\frac{\frac{\pi}{2}+\omega t}{2}$} in the equation above yields:
\begin{equation}
\small
\begin{aligned}
    \hat{x}_{sc}(t)
    &=\alpha_1 e^{\mu \frac{\frac{\pi}{2}+\omega t}{2}}\sum_{n=1}^{N}[r_{0}+ 2r_{2n}\cos(\frac{n \pi}{2}+n\omega t +\theta_{2n})]
    \\
    &~~~~
    +\alpha_2 e^{-\mu\frac{\frac{\pi}{2}+\omega t}{2}}\sum_{n=1}^{N}[r_{0}+ 2r_{2n}\cos(\frac{n\pi}{2}+n \omega t-\theta_{2n})].
    \label{Eq-Sol_r_th_2}
\end{aligned}
\end{equation}
Given initial condition ({\small$\hat{x}_{sc}(0)$}, {\small$\dot{\hat{x}}_{sc}(0)$}), we can compute the coefficients {\small$\alpha_1$} and {\small{$\alpha_2$}} based on the solution in \eqref{Eq-Sol_r_th_2}.}

\vspace{-0.4 in}
\begin{IEEEbiography}[{\includegraphics[width=1in,height=1.15in,clip,keepaspectratio]{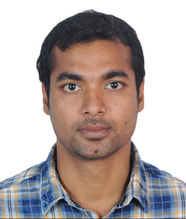}}]{Amir Iqbal} received a B.S. degree in Aerospace Engineering from the Indian Institute of Space Science and Technology, Thiruvananthapuram, Kerala, India, in 2012. In the past, he was a Scientist/Engineer at the ISRO Satellite Center, Bangalore, India. He is currently a Ph.D. candidate in the Department of Mechanical Engineering at the University of Massachusetts Lowell and a Research Intern at Purdue University.
\end{IEEEbiography}

\vspace{-0.5 in}
\begin{IEEEbiography}[{\includegraphics[width=1in,height=1.15in,clip,keepaspectratio]{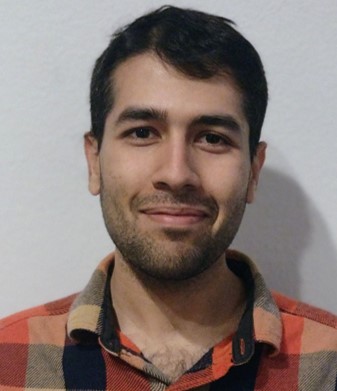}}]{Sushant Veer}
is a Senior Research Scientist at NVIDIA Research. In the past he was a Postdoctoral Research Associate in the Mechanical and Aerospace Engineering Department at Princeton University. He received his Ph.D. in Mechanical Engineering from the University of Delaware in 2018 and a B. Tech. in Mechanical Engineering from the Indian Institute of Technology Madras in 2013. His research interests lie at the intersection of control theory and machine learning with the goal of enabling safe decision making for robotic systems. He has received the Yeongchi Wu International Education Award (2013 International Society of Prosthetics and Orthotics World Congress), Singapore Technologies Scholarship (ST Engineering Pte Ltd), and Sri Chinmay Deodhar Prize (Indian Institute of Technology Madras).
\end{IEEEbiography}

\vspace{-0.4 in}
\begin{IEEEbiography}
[{\includegraphics[width=1in,height=1.15in,clip]{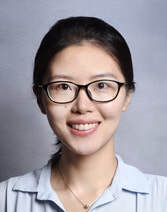}}]{Yan Gu}
received the B.S. degree in Mechanical Engineering from Zhejiang University, China, in June 2011 and the Ph.D. degree in Mechanical Engineering from Purdue University, West Lafayette, IN, USA, in August 2017.
She joined the faculty of the School of Mechanical Engineering at Purdue University in July 2022.
Prior to joining Purdue, she was an Assistant Professor with the Department of Mechanical Engineering at the University of Massachusetts Lowell.
Her research interests include nonlinear control, hybrid systems, legged locomotion, and wearable robots.
She was the recipient of the NSF CAREER Award in 2021.
\end{IEEEbiography}

\ifCLASSOPTIONcaptionsoff
  \newpage
\fi

\end{document}